\pdfoutput=1

\documentclass[11pt]{article}

\usepackage[]{ACL2023}

\usepackage{times}
\usepackage{latexsym}

\usepackage[T1]{fontenc}

\usepackage{times}
\usepackage{latexsym}
\usepackage{graphicx}
\usepackage{enumitem}
\usepackage{subfigure}
\usepackage{makecell}
\usepackage{amssymb}
\usepackage{lipsum}
\usepackage{booktabs}
\usepackage{multirow}
\usepackage{colortbl}
\usepackage{tabularx}
\usepackage{amsmath}
\usepackage{framed}
\usepackage{bbding}
\usepackage{lipsum}
\usepackage{longtable}
\usepackage{diagbox}
\usepackage[normalem]{ulem}
\useunder{\uline}{\ul}{}
\usepackage{CJKutf8}

\usepackage[utf8]{inputenc}
\usepackage{multirow}
\usepackage{cleveref}
\crefname{section}{§}{§§}
\Crefname{section}{§}{§§}
\usepackage[encapsulated]{CJK}

\usepackage{longtable}
\usepackage{booktabs}
\usepackage{graphicx}

\usepackage{xcolor}
\usepackage{microtype}

\usepackage{inconsolata}
\usepackage{arydshln}
\usepackage{wrapfig}
\usepackage{float}
\usepackage{listings}
\usepackage{fancyvrb}

\title{Large Language Model for Multi-Domain Translation: \\ Benchmarking and Domain CoT Fine-tuning}

\author{%
    Tianxiang Hu$^1$\thanks{\quad Tianxiang and Pei contributed equally. Work was done when Tianxiang was interning at Tongyi Lab.}\quad Pei Zhang$^{2,3}$$^*$\enspace Baosong Yang$^2$$^\dagger$\enspace Jun Xie$^2$\enspace Derek F. Wong$^3$\enspace Rui Wang$^1$\thanks{\quad Rui Wang and Baosong Yang are co-corresponding authors.}\\
    $^1$Shanghai Jiao Tong University\ \ \ $^2$Tongyi Lab\ \ \ $^3$NLP$^2$CT Lab, University of Macau\\
    $^1$\texttt{\small{\{hutianxiang,wangrui12\}@sjtu.edu.cn}} \\
    $^2$\texttt{\small{\{xiaoyi.zp,yangbaosong.ybs\}@alibaba-inc.com}}
}

\begin{document}
\maketitle
\begin{abstract}
Achieving consistent high-quality machine translation (MT) across diverse domains remains a significant challenge, primarily due to the limited and imbalanced parallel training data available in various domains. While large language models (LLMs) have demonstrated impressive general understanding and generation abilities, their potential in multi-domain MT is under-explored. We establish a comprehensive benchmark for multi-domain translation, featuring 25 German$\Leftrightarrow$English and 22 Chinese$\Leftrightarrow$English test sets respectively covering 15 domains. Our evaluation of prominent LLMs reveals a discernible performance gap against traditional MT systems, highlighting domain overfitting and catastrophic forgetting issues after fine-tuning on domain-limited corpora. To mitigate this, we propose a domain Chain of Thought (CoT) fine-tuning technique that utilizes the intrinsic multi-domain intelligence of LLMs to improve translation performance. This method inspires the LLM to perceive domain information from the source text, which then serves as a helpful hint to guide the translation process. Despite being trained on a small dataset of four domains, our CoT fine-tune approach achieves notable enhancements in translation accuracy and domain robustness than traditional fine-tuning, as evidenced by an average 1.53 BLEU score increase in over 20 German→English distinct out-of-domain tests.
\end{abstract}

\begin{CJK*}{UTF8}{gbsn}

\section{Introduction}
Significant progress has been made in the field of machine translation (MT) with the integration of deep learning~\citep{NIPS2017_3f5ee243}. Nonetheless, MT systems often display uneven translation quality when faced with diverse scenarios in real-world applications~\citep{koehn2017six,zhang-etal-2022-competency}. This inconsistency in translation quality is evident when comparing the competent performance in domains extensively seen during the training process against the inadequate performance in low-resource or unseen domains. 
This phenomenon highlights a critical multi-domain issue in MT~\citep{ currey-etal-2020-distilling, Chu2020, Pham2021}, where the translation ability of a model is hindered by the limited domain coverage and imbalanced distribution in parallel training corpora, leading to poor generalization across diverse domains. Despite this, acquiring a comprehensive and balanced training dataset that covers all domains is extremely challenging.

Generative large language models (LLMs) such as ChatGPT and GPT4~\citep{Achiam2023GPT4TR} have shown impressive general understanding and generation abilities across a broad range of tasks~\citep{Ouyang2022TrainingLM}. 
We believe LLMs pre-trained on massive amounts of corpora are inherently good at multi-domain understanding and generation, which could potentially address the dependence on parallel data for multi-domain MT. 
However, the potential of LLMs in multi-domain MT remains under-explored with most studies focusing on a narrow range of domains~\citep{currey-etal-2020-distilling, Man2023ExploringDA}. 
To comprehensively evaluate the translation capabilities while avoiding the inaccuracies caused by test leakage, we establish the most extensive multi-domain MT benchmark, which includes 25 German$\Leftrightarrow$English and 22 Chinese$\Leftrightarrow$English test sets covering 15 domains.

\begin{figure*}[bpht]
\begin{minipage}[t]{0.46\textwidth}
\centering
\includegraphics[width=1\textwidth]{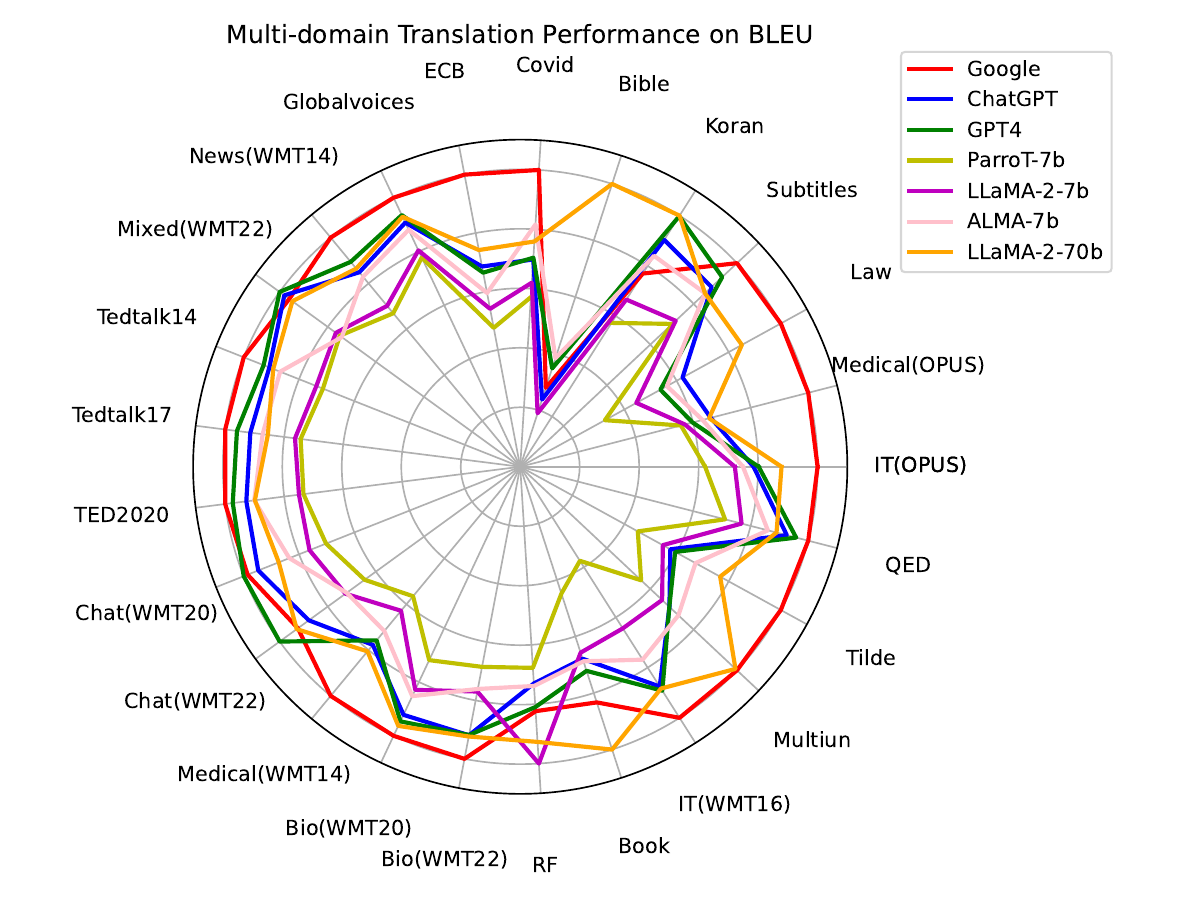}
\end{minipage}
\hfill
\begin{minipage}[t]{0.46\textwidth}
\centering
\includegraphics[width=1\textwidth]{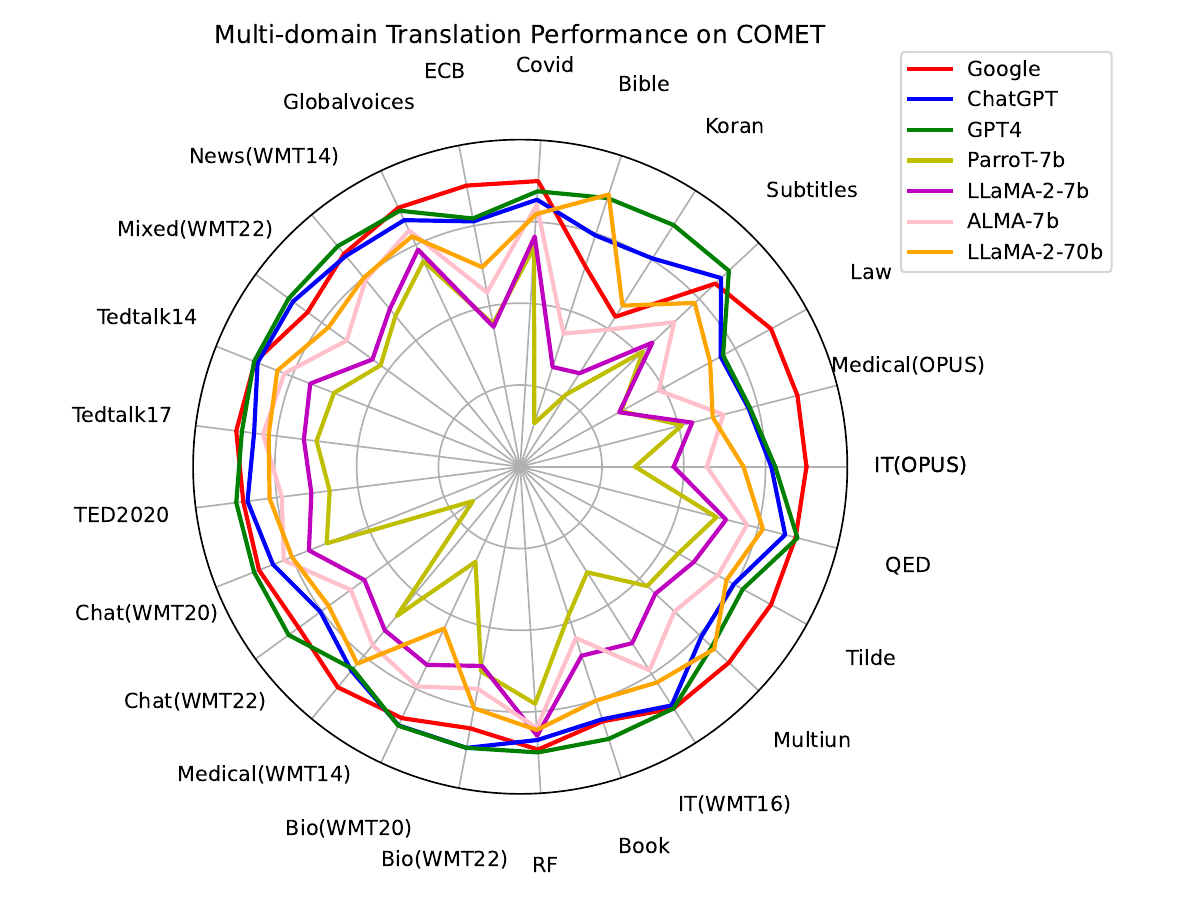}
\end{minipage}
 \caption{Performance comparision of prominent LLMs on the multi-domain German-to-English translation (other languages directions in Appendix \ref{sec:appendix_results}). For clear comparison, we show the scores normalized by the maximum score in each domain. The performance of LLMs varies greatly across multi-domains. Best reviewed in colors.}
\label{fig:radar}
\end{figure*}

By benchmarking prominent LLMs on four language directions, we find that LLMs, while promising, still exhibit obvious imbalanced performance across different domains (as illustrated in Figure \ref{fig:radar}). Recent efforts on LLMs such as prompt strategies~\citep{Wei2022ChainOT, li2022self, Zhang2023PromptingLL} and fine-tuning~\cite{Ouyang2022TrainingLM, Wang2022SelfInstructAL} have shown effective performance improvement for downstream tasks.
Nevertheless, in line with prior research~\cite{Thompson2019OvercomingCF, Lai2021ImprovingBD}, 
our study uncovers that when LLMs are fine-tuned on domain-specific parallel corpora (in-domain), their ability to generalize to unseen domains (out-of-domain) remains constrained, often resulting in issues such as catastrophic forgetting and overfitting~\cite{French1999CatastrophicFI, Saunders2021DomainAA}.
While certain methods like retrieval-based approaches~\cite{Agrawal2022IncontextES, Ghazvininejad2023DictionarybasedPP} and multi-perspective thinking~\cite{He2023ExploringHT} offer performance improvement, these techniques come at the cost of reliance on external knowledge resources and additional computational demands during decoding, aggravating the difficulty of practical deployment of LLMs.

In this study, our objective is to leverage the inherent multi-domain understanding and generative capacity of LLMs to improve multi-domain translation performance and enhance out-of-domain robustness. Our methodology is designed to inspire LLMs to elicit domain-specific insights from the source text, which are then used as a helpful hint prompt to guide the translation process. To accomplish this, we introduce a domain Chain-of-Thought (CoT) fine-tuning technique, which jointly cultivates LLM to recognize domain-specific information and output translation with given domain-specific hints during training. Despite being trained on only tens of thousands of parallel examples from four domains, our CoT fine-tuning method exhibits stronger translation quality and domain generalization than traditional fine-tuning. On the German$\Rightarrow$English translation task, it achieves an average 1.53 BLEU score improvement on 21 distinct out-of-domain tests and 0.83 average improvement on 4 in-domain tests.
Notably, this performance gain is magnified as the dataset is expanded to 400,000 examples and the model size is scaled to 70 billion.
Our approach has outperformed industry systems such as Google, GPT-4 and ChatGPT by exceeding 1.8 BLEU points averaged on the 25-domain benchmark, demonstrating its scalability and robustness when subjected to vastly larger data volumes and model capacity.

\section{Related Work}

\subsection{Multi-Domain Machine Translation}
Multi-domain machine translation is a crucial research topic with the goal of training a system to achieve robust translation performance across various domains~\cite{Pham2021, Man2023ExploringDA}. One approach focuses on studying domain adaptation~\cite{Saunders2021DomainAA, Thompson2019OvercomingCF, Lai2021ImprovingBD}, through fine-tuning to better adapt the model to new domains, but often results in catastrophic forgetting and overfitting issues. Another approach involves designing specialized domain knowledge modules, such as domain-specific and domain-shared knowledge learning mechanisms~\cite{Wang2019GoFT, lee-etal-2022-specializing, Man2023ExploringDA}. These tasks primarily rely on encoder-decoder neural machine translation. Our work focuses on addressing the multi-domain translation problem under LLMs and proposing new methods to resolve challenges in multi-domain translation.

\subsection{LLMs for Translation}
Recent research~\citep{Agrawal2022IncontextES, Moslem2023AdaptiveMT} has been focused on improving the translation capabilities of open-source LLMs through prompt strategies. ~\citet{Zhang2023PromptingLL} specifically investigate the impact of different templates on LLM for translation, while~\citet{Agrawal2022IncontextES} investigates example selection strategies in in-context learning for LLM. ~\citet{Moslem2023AdaptiveMT} explores incorporating external knowledge such as dictionaries or phrases to aid in translation, and~\citet{He2023ExploringHT} design a multi-dimensional prompt strategy inspired by human thinking. Additionally, ~\citet{Raunak2023LeveragingGF} and~\citet{Koneru2023ContextualRO} utilize LLMs for post-editing or refining translation outputs. However, these methods either heavily rely on external knowledge or exhibit subpar performance on LLMs of the scale of 7B.

Fine-tuning is another common method to enhance the performance of LLM by improving their ability to follow NLP task instructions~\cite{Ouyang2022TrainingLM, Wang2022SelfInstructAL, alpaca}. Significant performance improvements can be achieved by fine-tuning with specific translation task data~\cite{Jiao2023ParroTTD}. Previous work has developed various fine-tuning datasets. ALMA~\cite{Xu2023APS} uses monolingual data to improve low-resource languages, ParroT~\cite{Jiao2023ParroTTD} combines human-written translation and feedback data to guide high-quality translation, TIM~\cite{Zeng2023TIMTL} uses examples in comparison data to teach models what is better translation. These approaches mainly enhance translation quality or multilingual translation abilities by constructing specific instruction data.

\section{Benchmarking LLMs for Multi-domain Machine Translation}

The currently available multi-domain benchmarks suffer from deficient domain coverage, and the publicly available test sets have potential risks of data leakage in the MT training process, leading to skewed performance evaluations.
Moreover, as MT systems exhibit advancing performance and generalization capabilities, there lacks a finer-grained and more precise evaluation to keep pace with the rapid advancements in multi-domain MT.

To provide a thorough and precise evaluation, we construct the most comprehensive multi-domain MT benchmark for two language pairs, primarily sourced from OPUS, WMT, TedTalks, in-house tests and prior works~\citep{Aharoni2020UnsupervisedDC, Tian2014UMCorpusAL}. There are 25 tests for German-English (De$\Leftrightarrow$En) and 22 for Chinese-English (Zh$\Leftrightarrow$En), each spanning 15 distinct domains such as news, Global Voices, COVID, medical, industry, lecture, subtitle, chat, book, law, government, finance, IT, religion, science, novel, energy.~\footnote{The specific domains covered by De$\Leftrightarrow$En and Zh$\Leftrightarrow$En pairs differ, and details can be found in the Appendix~\ref{sec:appendix_data}.}

\paragraph{Settings}

We conducted experiments on bidirectional MT tasks for De$\Leftrightarrow$En and Zh$\Leftrightarrow$En.
To better evaluate translation performance, we adopt two widely-used metrics: SacreBLEU~\cite{Papineni2002BleuAM,Post2018ACF}, a n-gram matching-based metric, and COMET\footnote{We use Unbabel/wmt22-comet-da for COMET.}~\cite{Rei2020COMETAN}, a supervised model that based on pre-trained language models.

We evaluate seven leading LLMs, using tailored few-shot inference strategies to suit the unique features of each LLM.
For LLaMA-2-7b, which isn't trained to follow instructions, we use a 5-shot approach. For models with instruction following ability, such as LLaMA-2-70b, GPT-4~\footnote{Due to budget constraints, we only evaluate GPT-4 on De$\rightarrow$En translation task.} and ChatGPT, we employ a 1-shot strategy. For models specifically trained for MT, like ALMA-7b~\cite{Xu2023APS}, BayLing-7b~\cite{bayling}, and ParroT-7b~\cite{Jiao2023ParroTTD}, we apply a 0-shot method. The details of prompt templates are shown in Appendix~\ref{sec:appendix_prompt}. We also present the commercial general MT Google Translate as a reference.

\paragraph{Results}

\begin{table*}[h]
\centering
\resizebox{0.85\textwidth}{!}{
\begin{tabular}{lccccccccc}
\toprule
\multirow{2}{*}{\textbf{Method}} & \multicolumn{2}{c}{\textbf{De}$\Rightarrow$\textbf{En}}&\multicolumn{2}{c}{\textbf{En}$\Rightarrow$\textbf{De}}&\multicolumn{2}{c}{\textbf{Zh}$\Rightarrow$\textbf{En}}&\multicolumn{2}{c}{\textbf{En}$\Rightarrow$\textbf{Zh}}&\multicolumn{1}{c}{\textbf{Avg.}}\\
&In-domain&OOD&In-domain&OOD&In-domain&OOD&In-domain&OOD&\\
\hline
&\multicolumn{9}{c}{BLEU} \\
        Google & 41.63 & 39.51 & 35.23 & 32.72 & 27.54 & 31.84 & 41.15 & 44.90 & \textbf{37.06}  \\ 
        ChatGPT & 35.84 & 36.75 & 35.84 & 29.75 & 35.84 & 27.95 & 35.84 & 38.16 & 32.82  \\ 
        \hdashline
        ParroT-7b  & 31.95 & 32.35 & 24.10 & 22.97 & 18.14 & 20.45 & 26.13 & 23.97 & 24.96  \\ 
        BayLing-7b  & 33.13 & 33.71 & 24.79 & 23.05 & 19.35 & 22.55 & 28.92 & 30.59 & 27.30  \\
        LLaMA-2-7b   & 33.30 & 34.34 & 25.31 & 23.13 & 19.72 & 23.26 & 26.05 & 26.21 & 26.61  \\ 
        ALMA-7b  & 35.02 & 35.85 & 28.72 & 26.47 & 22.56 & 25.36 & 31.14 & 33.35 & 30.09  \\ 
        LLaMA-2-70b  & 37.48 & 38.58 & 30.89 & 28.66 & 26.68 & 28.31 & 34.34 & 35.20 & 32.62  \\ 
        \hdashline
        LLaMA-2-7b   & ~ & ~ & ~ & ~ & ~ & ~ & ~ & ~ &   \\
                + BM25  & 36.90 & 34.23 & 28.47 & 22.86 & 23.78 & 22.34 & 28.43 & 24.82 & 26.62  \\ 
                + MAPS  & 33.73 & 34.28 & 25.37 & 22.32 & 20.07 & 22.52 & 25.88 & 24.84 & 26.02  \\ 
                + FT   & 40.16 & 35.36 & 32.23 & 24.51 & 25.98 & 21.39 & 30.43 & 27.86 & 28.10  \\ 
                + FT + BM25  & 41.59 & 36.01 & 32.66 & 23.96 & 27.40 & 21.65 & 31.40 & 26.42 & 28.07  \\ 
                + FT + MAPS  & 39.58 & 36.68 & 29.66 & 23.35 & 21.41 & 22.35 & 27.81 & 26.27 & 27.56 \\ 
\hline\hline
&\multicolumn{9}{c}{COMET} \\
        Google & 83.86 & 85.91 & 83.25 & 85.60 & 84.37 & 82.38 & 86.80 & 87.68 & \textbf{85.27}  \\ 
        ChatGPT & 83.16 & 85.77 & 81.99 & 84.83 & 83.51 & 82.89 & 85.51 & 86.48 & 84.76  \\ 
        \hdashline
        ParroT-7b  & 80.98 & 83.72 & 77.74 & 80.40 & 79.56 & 77.07 & 80.33 & 79.08 & 80.03  \\ 
        BayLing-7b  & 81.25 & 84.03 & 78.30 & 80.71 & 81.14 & 79.04 & 82.68 & 82.77 & 81.52  \\ 
        LLaMA-2-7b   & 81.28 & 84.39 & 77.94 & 81.34 & 81.09 & 79.00 & 81.31 & 81.45 & 81.37  \\ 
        ALMA-7b  & 82.00 & 84.92 & 80.71 & 83.40 & 82.49 & 80.71 & 84.16 & 84.81 & 83.29  \\ 
        LLaMA-2-70b  & 82.58 & 85.29 & 81.00 & 83.37 & 83.22 & 81.37 & 84.51 & 84.42 & 83.50  \\ 
        \hdashline
        LLaMA-2-7b   & ~ & ~ & ~ & ~ & ~ & ~ & ~ & ~ &   \\ 
                + BM25  & 81.69 & 83.72 & 78.22 & 80.52 & 81.47 & 78.40 & 80.22 & 79.66 & 80.57  \\ 
                + MAPS  & 82.26 & 85.08 & 79.04 & 81.61 & 82.14 & 79.94 & 81.66 & 81.07 & 81.83  \\ 
                + FT   & 83.63 & 84.32 & 81.60 & 82.29 & 82.53 & 78.63 & 83.09 & 82.70 & 82.13  \\
                + FT + BM25  & 83.48 & 84.56 & 80.76 & 81.81 & 82.23 & 78.34 & 82.77 & 81.81 & 81.76  \\ 
                + FT + MAPS  & 83.73 & 85.51 & 80.77 & 82.18 & 82.47 & 80.09 & 82.80 & 82.27 & 82.52 \\
\bottomrule
\end{tabular}
}
\caption{Evaluation of different methods on multi-domain test sets for four language pairs. For the results of BLEU and COMET, the top is the commercial system, the middle is open-source LLMs and fine-tuning (FT) LLMs, and the bottom is some enhancements based on LLaMA2-7b. In domain refers to the four training domains for fine-tuning, while out-of-domain (OOD) refers to other domains. LLMs in all four language directions still have a certain gap compared to Google. Fine-tuning can significantly improve in-domain translation performance, while there may be a decline in some OOD domains.}
 \label{tab:fine-tuning}
\end{table*}

As shown in Figure~\ref{fig:radar}, Google outperforms others and demonstrates a more balanced performance across various domains. While ChatGPT and GPT-4 do not reach the highest BLEU scores, they are on par with Google in the COMET metric, especially in tasks where the target language is English. Due to the limitations in model capacity, open-source LLMs such as BLOOMZ-7b and LLaMA-2-7b significantly lag behind their commercial counterparts. Scaling up models such as LLaMA-2-70b yields certain improvements.
Models specially fine-tuned with translation instructions, such as ParroT, Bayling and ALMA, display more obvious imbalances across various domains. This underscores a persistent challenge in multi-domain translation that has not been adequately addressed in prior research.
Models of the LLM community commonly exhibit issues of performance imbalance. There is a substantial deficiency in professional domains such as Law, Medical, IT, and ECB, while the gap is relatively smaller in oral communication domains such as Chat.
These findings highlight the necessity for research on how to stimulate translation capabilities and domain robustness of LLMs.

\section{Analyzing LLM’s Translation Performance with Fining-Tuning}

To evaluate the efficacy of fine-tuning LLMs for translation tasks, we follow prior work~\citep{Jiao2023ParroTTD} to construct a translation instruction dataset in the Stanford Alpaca~\citep{alpaca} format for the fine-tuning process.

\paragraph{Settings}
We conduct experiments with DeepSpeed-Chat\footnote{\url{https://github.com/microsoft/DeepSpeedExamples}} and use LLaMA-2-7b as the backbone. We fine-tune LLaMA-2-7b with a batch size of 16, a learning rate of 1e-4, a weight decay of 0.1, and a maximum text length of 512 tokens. We use LoRA~\citep{Hu2021LoRALA} for fine-tuning, with the LoRA dim set to 8 and the LoRA dropout set to 0.1. We conduct fine-tuning on two NVIDIA A100 GPUs and utilize DeepSpeed ZeRO stage 3 for model parallelism.

For De$\Leftrightarrow$En training, we use ~\citet{Aharoni2020UnsupervisedDC}'s multi-domain training data set and randomly sample 10k training data from each domain of Medical, Law, IT, and Subtitles.
For Zh$\Leftrightarrow$En training, we use UM-Corpus~\citep{Tian2014UMCorpusAL} multi-domain training data set and randomly sample 10k training data from each domain of News, Science, Laws, and Subtitles. To ensure fair evaluation, we select the model after fine-tuning for one epoch during inference. 

\begin{figure*}[h]
  \centering
  \includegraphics[width=\textwidth]{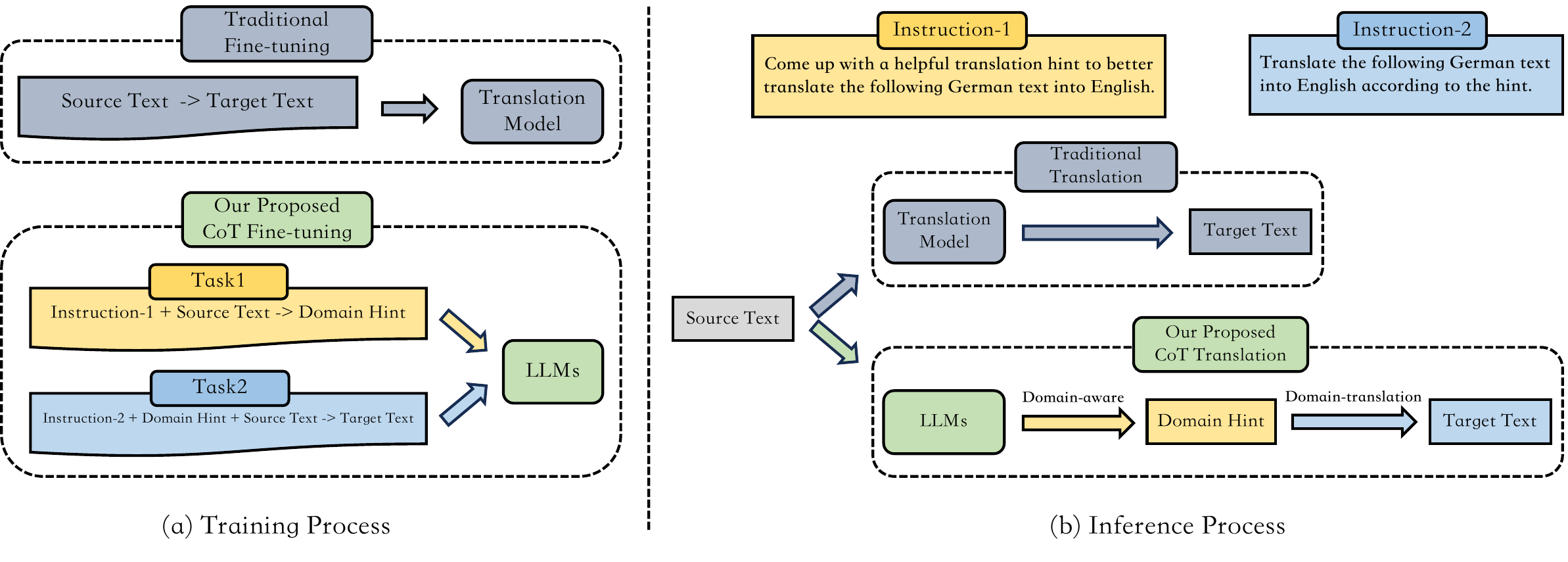}
  \caption{Comparison of our proposed CoT fine-tuning and traditional fine-tuning framework in training and inference process. During the training process, it includes a domain hint generation task and a domain translation task; During the inference process, it first generates domain-aware hints and then undertakes domain translation based on the generated domain hints.}
  \label{fig:CoT}
\end{figure*}

\paragraph{Results}
We utilize LLaMA-2-7b as the backbone and evaluate the following baselines:

\textbf{1) BM25}~\cite{Agrawal2022IncontextES}: which retrieves the most similar example to the test source from the training datastore via the BM25. We use a 1-shot example for translation. \textbf{2) MAPS}~\cite{He2023ExploringHT}: which designs a multi-dimensional prompt strategy inspired by human thinking and then uses QE scorer to select the best translation. We used wmt21-comet-qe-da as the QE scorer based on the original paper. \textbf{3) FT}: which fine-tunes the multi-domain training data. \textbf{4) FT + BM25}: which first fine-tunes the base model and then uses BM25 strategy on the fine-tuned model. \textbf{5) FT + MAPS}: which first fine-tunes the base model and then uses MAPS strategy on the fine-tuned model.

The results present in Table \ref{tab:fine-tuning} indicate that BM25 and MAPS do not result in a significant improvement in BLEU and COMET metrics compared to the base model LLaMA-2-7b. This suggests that these methods have limitations in the context of multi-domain translation. Conversely, the fine-tuning approach demonstrated a noticeable improvement in in-domain performance and a slight improvement in out-of-domain performance. Even when BM25 retrieval and MAPS strategies are applied to the fine-tuning model, there is no further significant performance improvement compared to the model with only fine-tuning, and in some cases, the performance even decreases. This may be attributed to the decreased in-context learning capability of the model after fine-tuning~\citep{Alves2023SteeringLL}. The findings suggest that fine-tuning LLM to a specific task using supervised data can be an effective approach.

\paragraph{Overfitting Problem in Fine-Tuning LLM}
We also observe catastrophic forgetting and overfitting issues. Based on the comprehensive translation performance of each domain presented in the Appendix \ref{sec:appendix_results}, we can see that the performance improvement is notable only within the in-domain, with minimal impact on the most unseen domains, and in some cases, there is no improvement or even a decline. We further analyze this phenomenon. As shown in Figure \ref{fig:overfitting}, it is apparent that fine-tuning LLM is susceptible to overfitting. As the training epoch progresses, the performance within the in-domain continues to improve and then stabilizes or slightly decreases, while the performance within the OOD initially exhibits a slight improvement, followed by a significant decline.

\section{Method}

\subsection{Chain-of-Thought Fine-tuning}
Our research focuses on the challenge of multi-domain translation. Simply fine-tuning parallel bilingual translation data can lead to the problem of overfitting during training, resulting in poor performance on OOD data. To address this issue, we propose a fine-tuning approach based on hints, aiming to leverage the internal knowledge of the LLM itself without introducing external knowledge. This enables the model to self-perceive, comprehend the input sentence, and generate the corresponding domain hint. This approach allows us to activate the existing knowledge of LLM and generate translations with different domain styles based on different domain features. And using different translation hint templates for different domains helps us avoid overfitting to domain-specific templates when translating OOD data.

Figure \ref{fig:CoT} illustrates the framework of our proposed CoT fine-tuning (CoT-FT) training strategy and inference stage. The training process involves two tasks: domain hint prediction and generation, followed by translation based on the generated domain hint. We use two sets of instructions to distinguish between these tasks. To enable the model to learn to recognize different domains and generate corresponding translations based on the given hints, we introduce the "Hint" section in the domain translation task. This section mainly describes the sentence's domain style and other characteristics to be translated (for detailed data prompt, please refer to the Appendix \ref{sec:appendix_sft}). In the inference process, the model first generates domain-aware hints according to Instrcution-1 and then performs domain translation based on the generated domain hints according to Instrcution-2, as shown in Figure \ref{fig:CoT}.

By jointly training the model to generate domain hints and perform multi-domain translation tasks, we enable LLM to learn both domain discrimination and multi-domain translation capabilities across multiple domains. The loss function for the domain generation task is defined as follows:
\setlength{\abovedisplayskip}{5pt} 
\setlength{\belowdisplayskip}{5pt} 
\begin{align}
    \mathcal{L}_{\text{hint}}(\boldsymbol{\theta})&=-\log P(\mathbf{h}|\mathrm{inst_1},\mathbf{x};\boldsymbol{\theta})\\
           &=-\sum_{t=1}^{T_1}\log P(h_t|\mathrm{inst}_1,\mathbf{x},\mathbf{h}_{\textless t};\boldsymbol{\theta}),
\end{align}
where $\mathbf{x}$ is the monolingual source text, $\mathbf{h}$ is the domain hint of $\mathbf{x}$, $\mathrm{inst}_1$ is the instruction for generating domain hint, $T_1$ is the length of $\mathbf{h}$ and $\boldsymbol{\theta}$ is the model parameters. The domain translation loss can be formulated as follows:
\begin{align}
    \mathcal{L}_{\text{trans}}(\boldsymbol{\theta})&=-\log P(\mathbf{y}|\mathrm{inst}_2,\mathbf{h},\mathbf{x};\boldsymbol{\theta})\\
           &=-\sum_{t=1}^{T_2}\log P(y_t|\mathrm{inst}_2,\mathbf{h},\mathbf{x},\mathbf{y}_{\textless t};\boldsymbol{\theta}),
\end{align}
where $\mathbf{y}$ is the target translation text, $\mathrm{inst}_2$ is the instruction for domain translation, and $T_2$ is the length of $\mathbf{y}$. Then we use $\alpha$ to control the proportion of data between two tasks, and the final loss function is:
\begin{align}
    \mathcal{L}_{\text{cot\_sft}}&=\mathcal{L}_{\text{trans}}+\alpha \mathcal{L}_{\text{hint}}.
\end{align}

\begin{table*}[htbp]
\centering
\resizebox{0.9\textwidth}{!}{
\begin{tabular}{lcccccccccc}
\toprule
\multirow{2}{*}{\textbf{Method}} & \multicolumn{2}{c}{\textbf{De}$\Rightarrow$\textbf{En}}&\multicolumn{2}{c}{\textbf{En}$\Rightarrow$\textbf{De}}&\multicolumn{2}{c}{\textbf{Zh}$\Rightarrow$\textbf{En}}&\multicolumn{2}{c}{\textbf{En}$\Rightarrow$\textbf{Zh}}&\multicolumn{1}{c}{\textbf{Avg.}}\\
&In-domain&OOD&In-domain&OOD&In-domain&OOD&In-domain&OOD&\\
\hline
LLaMA-2-7b & \multicolumn{8}{c}{BLEU} \\ 
\hdashline
        + FT   & 40.16 & 35.36 & 32.23 & 24.51 & 25.98 & 21.39 & 30.43 & 27.86 & 28.10  \\ 
        + FT + BM25  & \textbf{41.59} & 36.01 & \textbf{32.66} & 23.96 & \textbf{27.40} & 21.65 & \textbf{31.40} & 26.42 & 28.07  \\ 
        + FT + MAPS  & 39.58 & 36.68 & 29.66 & 23.35 & 21.41 & 22.35 & 27.81 & 26.27 & 27.56  \\ 
\hdashline
        + CoT-FT & 40.99 & 36.99 & 32.47 & 24.65 & 26.41 & 22.44 & 30.02 & 27.60 & 28.68  \\
        + CoT-FT-R & 39.04 & 36.45 & 32.12 & 24.39 & 25.53 & 22.15 & 30.10 & 27.69 & 28.35  \\ 
        + CoT-FT-G & 41.16 & \textbf{37.67}& 32.50 & \textbf{24.87} & 26.57 & \textbf{23.12} & 29.65 & \textbf{28.17} & \textbf{29.13} \\ 
\hline\hline
LLaMA-2-7b & \multicolumn{8}{c}{COMET} \\ 
\hdashline
        + FT   & 83.63 & 84.32 & 81.60 & 82.29 & 82.53 & 78.63 & \textbf{83.09} & 82.70 & 82.13  \\ 
        + FT + BM25  & 83.48 & 84.56 & 80.76 & 81.81 & 82.23 & 78.34 & 82.77 & 81.81 & 81.76  \\ 
        + FT + MAPS  & 83.73 & \textbf{85.51} & 80.77 & 82.18 & 82.47 & \textbf{80.09} & 82.80 & 82.27 & 82.52  \\ 
\hdashline
        + CoT-FT & 83.91 & 84.87 & 81.62 & 82.44 & 82.57 & 79.44 & 82.57 & 82.44 & 82.37  \\ 
        + CoT-FT-R & 83.12 & 84.73 & 81.25 & 82.30 & 82.43 & 79.32 & 82.62 & 82.51 & 82.25 \\  
        + CoT-FT-G & \textbf{84.05} & 85.15 & \textbf{81.67} & \textbf{82.54} & \textbf{82.62} & 79.99 & 82.33 & \textbf{82.79} & \textbf{82.64} \\ 
\bottomrule
\end{tabular}
}
\caption{Evaluation of different methods on four language pairs for multi-domain test sets. Bold entries denote statistically significant differences with p < 0.05 in the paired t-test compared to other methods. Our approach demonstrates consistent improvements in both BLEU and COMET metrics. $^*$ MAPS uses COMET21 as a selector and may suffer the problem of overfitting evaluation metrics, leading to strong performance in COMET22.}
 \label{tab:main}
\end{table*}

\subsection{Experiments}

\paragraph{Settings}
Unless otherwise specified, the training and decoding settings remain consistent with those outlined in the previous section on fine-tuning. For CoT fine-tuning, we use 1/10 of the domain translation task data for the domain generation task. The domain generation task data is equally distributed across ten domains for De$\Leftrightarrow$En and eight domains for Zh$\Leftrightarrow$En, with a manually crafted domain-style-related prompt hint for each domain. Since the monolingual data is relatively easier to collect compared to domain-specific bilingual data, the acquisition cost is significantly lower than that of bilingual data. Therefore, we do not think it introduces much unfairness compared to other methods. Please refer to the Appendix \ref{sec:appendix_sft} for more specific details.
\paragraph{Baselines}

\textbf{1)} FT, FT + BM25 and FT + MAPS are the experimental results of the third section. 
\textbf{2)} CoT-FT: which uses the proposed strategy.
\textbf{3)} CoT-FT-G: which uses the proposed strategy. Specifically, we manually provide a domain hint for each domain during decoding. When translating a sentence, we give the actual domain hint of that sentence before proceeding with the second translation step.
\textbf{4)} CoT-FT-R: which uses the proposed strategy. Specifically, we provide a random hint from the 25 domain hints we create in CoT-FT-G during decoding.

\paragraph{Main Results}
As indicated in Table \ref{tab:main}, the CoT-FT-G model demonstrates superior performance in both BLEU and COMET metrics. Despite not receiving explicit domain hints, the CoT-FT model can autonomously generate relevant domain hints for input sentences, resulting in significantly improved BLEU scores compared to BM25 and MAPS, and slightly lower COMET scores than MAPS. Overall, CoT-FT outperforms the other models on average across multiple metrics. BM25 performs well in BLEU but not in COMET, while MAPS, using COMET21 as a selector, may suffer the problem of overfitting evaluation metrics, leading to strong performance in COMET22 but a notable reduction in BLEU scores. Additionally, our proposed method outperforms direct fine-tuning in De-En, En-De, and Zh-En translation tasks, particularly for OOD test sets. This suggests the effectiveness of CoT fine-tuning strategy in enabling LLMs to discern and generate relevant domain hints, ultimately enhancing translation performance. It should be noted that due to the in-domain and OOD datasets having different test sets, a higher COMET score on OOD than in-domain does not prove that the translation quality on out-of-domain is superior. 

We also observe that our proposed methods significantly impact X$\Rightarrow$En translation, while En$\Rightarrow$X  exhibits a weaker performance and even a decreased performance sometimes(En$\Rightarrow$Zh). We speculate that LLaMA-2 is primarily pre-trained on English corpora, containing substantial English knowledge. Therefore, specific hints can activate LLM's inherent knowledge when generating English translations, improving performance. However, LLaMA-2 has less Chinese knowledge, the generated hints may introduce noise, resulting in performance degradation.

\paragraph{Human Evaluation}
We randomly selected 110 sentences for human quality assessment in Zh$\Rightarrow$En translation tasks. Our professional annotators were presented with a source sentence and two translations. They were asked to select which translation was superior from the perspectives of accuracy, naturalness, fluency, consistency, linguistic style, and cultural adaptability. Table \ref{tab:human evaluation} shows that our CoT-FT outperforms the FT approach.
\begin{table}[htbp]
\centering
\resizebox{0.45\textwidth}{!}{
\begin{tabular}{l|ccc}
\toprule
Human Evaluation&CoT-FT Lose&CoT-FT Win&Tie\\
\hline
En-Zh CoT-FT vs. FT&22.7\%&31.8\%&45.5\% \\
Zh-En CoT-FT vs. FT&20.0\%&32.7\%&47.3\% \\
\bottomrule
\end{tabular}
}
\caption{Human evaluation study, comparing CoT-FT with direct FT.}
 \label{tab:human evaluation}
\end{table}

\paragraph{Ablation Study}
Table \ref{tab:main} illustrates the performance gap among the self-domain hints produced by our CoT fine-tuning method, randomly selected domain hints, and human-defined domain hints. The domain-aware hints outperform randomly selected hints, although they slightly underperform compared to the provided domain hints. This suggests that our proposed strategy, without introducing external knowledge, enhancing the performance of multi-domain machine translation.

\subsection{Analysis}

\begin{figure}[htpb]
  \centering
  \includegraphics[width=0.45\textwidth]{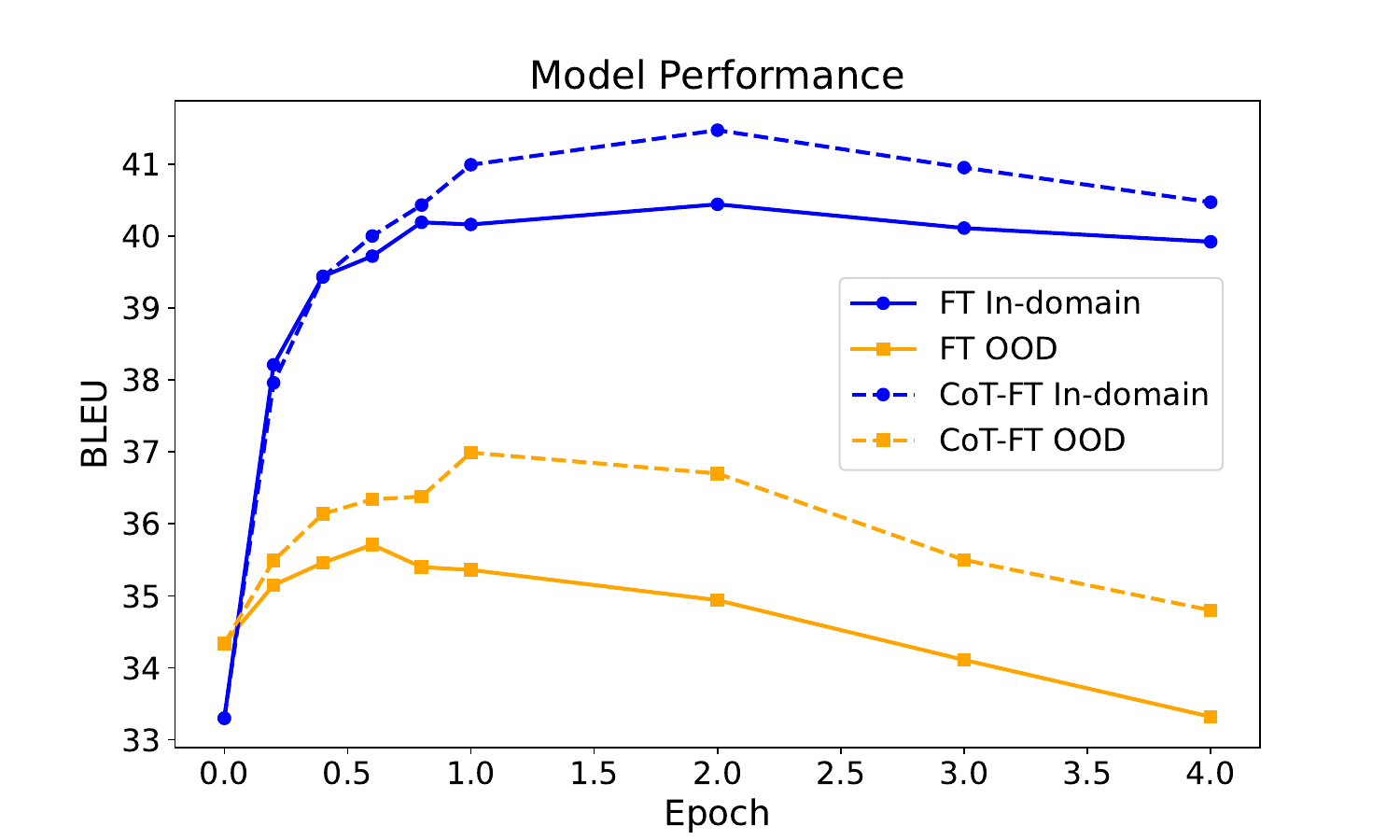}
  \includegraphics[width=0.45\textwidth]{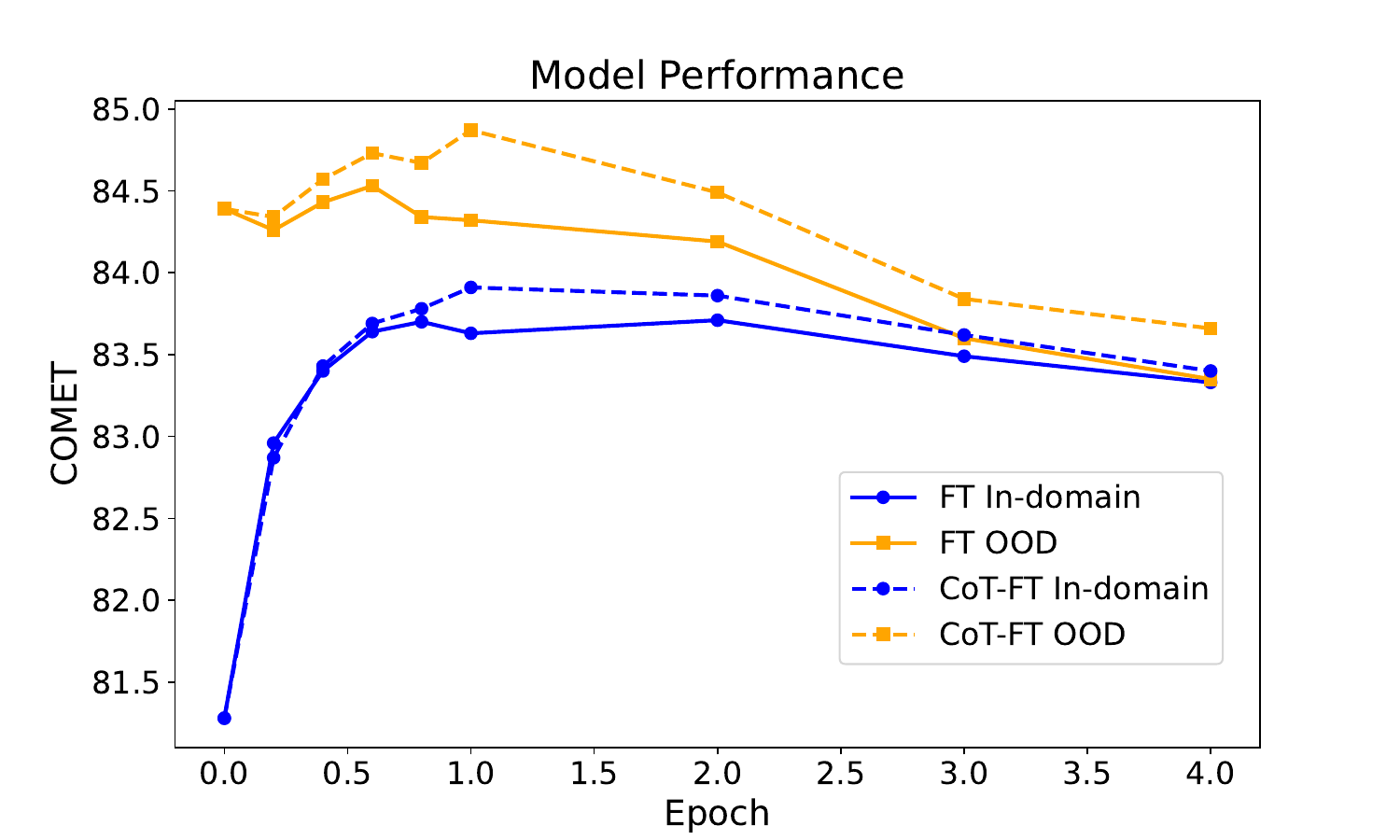}
  \caption{Model performance changes as training epoch progresses. For De-En, with fine-tuning on LLaMA-2-7b, there is an overfitting problem on OOD data. CoT-FT alleviates the problem to a certain extent.}
  \label{fig:overfitting}
\end{figure}

\paragraph{Alleviating Overfitting Phenomenon}
Figure \ref{fig:overfitting} illustrates that our proposed CoT fine-tuning strategy outperforms direct fine-tuning as the number of epochs increases, both in-domain and OOD scenarios. Our approach effectively improves performance on OOD during the early stages of training, mitigating overfitting and enhancing the overall performance and generalization capability of the model for multi-domain MT.

\begin{table}[htbp]
\centering
\resizebox{0.45\textwidth}{!}{
\begin{tabular}{l|cc}
\toprule
\textbf{Method}& BLEU in-domain & BLEU OOD \\
\hline
FT & 40.16 & 35.36 \\
Hint-FT-G & 40.85 & 37.23 \\
CoT-FT & 40.99 & 36.99 \\
CoT-FT-G & 41.16 & 37.67 \\
\bottomrule
\end{tabular}
}
\caption{The effect of introducing the hint loss.}
 \label{tab:hint loss}
\end{table}

\paragraph{Effects of the hint loss}
We employ the Hint-FT-G strategy (instruction + hint + source -> target), meaning during training we do not introduce a hint loss, while during decoding manual hints are provided. From Table \ref{tab:hint loss}, comparing Hint-FT-G with CoT-FT-G, it can be seen that introducing hint loss can further enhance performance.

\begin{figure}[h]
  \centering
  \includegraphics[width=0.45\textwidth]{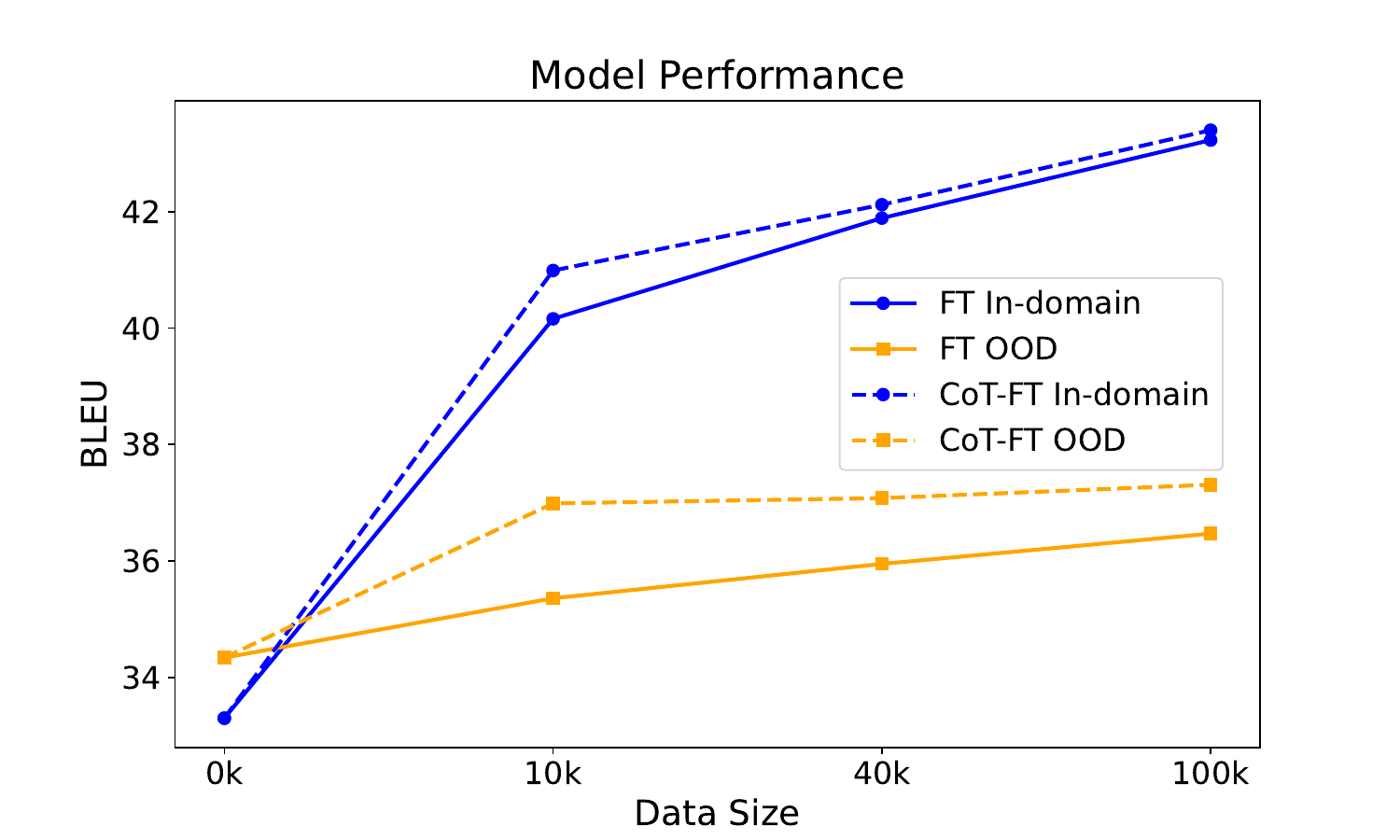}
  \includegraphics[width=0.45\textwidth]{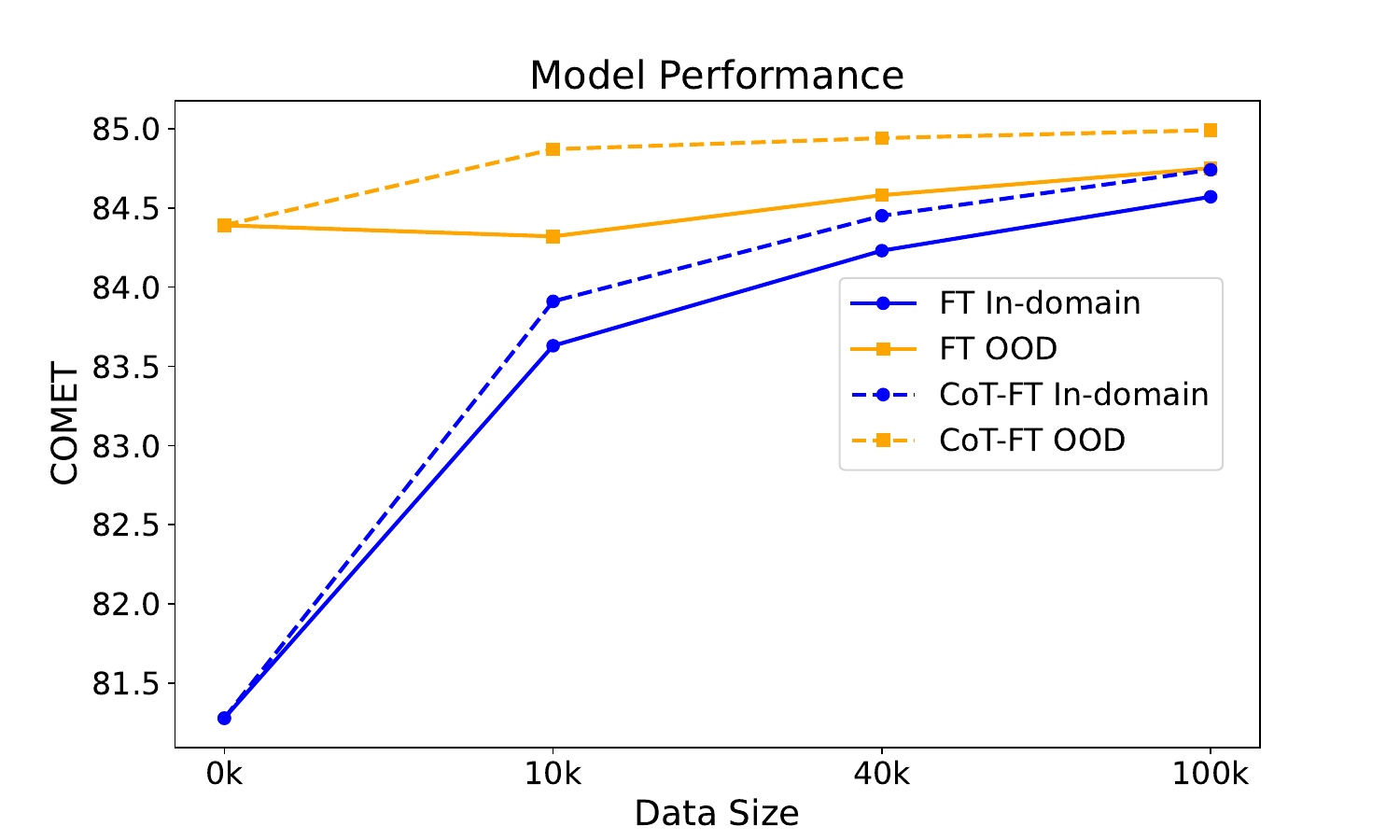}
  \caption{BLEU and COMET scores for FT and CoT-FT at different data sizes(De-En). The proposed CoT fine-tuning outperforms direct fine-tuning in both in-domain and OOD scenarios.}
  \label{fig:data scaling}
\end{figure}

\paragraph{Effects of Data Size Scaling}
We examine the impact of varying training data sizes on model performance. As shown in Figure \ref{fig:data scaling}, with the training data size increasing, the model's performance shows notable improvement in the early stage. However, further increasing the data size does not yield significant performance enhancements to the same extent. CoT fine-tuning demonstrates significantly superior performance in both in-domain and OOD settings compared to direct fine-tuning with training data of different sizes. Particularly in the OOD scenario, there is a marked improvement in BLEU and COMET scores.

\paragraph{Effects of Model Size Scaling}
In this section, we provide a comparative analysis of direct fine-tuning and CoT fine-tuning across various model sizes on the multi-domain test sets. The LLaMA2 model comes in sizes 7b, 13b, and 70b, with a training data size of 4*100k. Table \ref{tab:model size scaling} illustrates that our method can achieve stable performance improvement on both in domain and OOD under different model size scaling, indicating its generalizability. The translation performance of the LLMs fine-tuned with CoT-FT improves as the foundational LLM's size increases. Moreover, when using LLaMA-2-70b as the base model, as shown in Table \ref{tab:model size scaling 2}, our method demonstrates an average performance that exceeds Google Translator and ChatGPT in both BLEU and COMET metrics.

\begin{table}[htbp]
\centering
\resizebox{0.45\textwidth}{!}{
\begin{tabular}{l|cccc}
\toprule
\multirow{2}{*}{\textbf{Method}} & \multicolumn{2}{c}{In-domain}&\multicolumn{2}{c}{OOD}\\
&BLEU&COMET&BLEU&COMET\\
\hline
        LLaMA-2-7b & 33.30 & 81.28 & 34.34 & 84.39   \\
        + FT & 43.23 & 84.57 & 36.67 & 84.75 \\  
        + CoT-FT & \textbf{43.46} & \textbf{84.74}  & \textbf{37.31} & \textbf{84.99}  \\
        \hdashline
        LLaMA-2-13b & 34.70 & 81.55 & 35.45 & 84.64 \\ 
        + FT & 44.31 & 84.76 & 37.69 & 84.93 \\ 
        + CoT-FT & \textbf{44.31} & \textbf{84.95} & \textbf{38.36} & \textbf{85.31} \\ 
        \hdashline
        LLaMA-2-70b & 37.48 & 82.58 & 38.58 & 85.29 \\
        + FT & 46.28 & 85.25 & 39.78 & 85.54 \\
        + CoT-FT & \textbf{46.46} & \textbf{85.41} & \textbf{40.72} & \textbf{85.74} \\
\bottomrule
\end{tabular}
}
\caption{Performance comparison of different methods at 7b, 13b, 70b model sizes (with a training data size of 4*100k). CoT fine-tuning significantly outperforms direct fine-tuning across different model sizes.}
 \label{tab:model size scaling}
\end{table}

\begin{table}[htbp]
\centering
\resizebox{0.45\textwidth}{!}{
\begin{tabular}{l|cc}
\toprule
\textbf{Method}& BLEU & COMET \\
\hline
        Google & 39.85 &  85.58 \\
        ChatGPT & 36.61 &  85.35  \\
        GPT4 & 37.11 &  85.60  \\
        LLaMA-2-70b + CoT-FT & \textbf{41.64} & \textbf{85.69}  \\
\bottomrule
\end{tabular}
}
\caption{Comparison of average performance on 25 German$\Rightarrow$English domain test sets.}
 \label{tab:model size scaling 2}
\end{table}
\section{Conclusion}
In this paper, we present a comprehensive study of LLMs for multi-domain MT. Our study begins by constructing a multi-domain MT testing benchmark and evaluating the performance of popular LLMs. Our findings reveal that multi-domain MT remains a challenging task on LLMs. Then, we conduct an analysis of previous approaches, noting that many of them have not fully addressed the challenges of multi-domain MT. Finally, we propose a domain Chain of Thought fine-tuning strategy, which involves jointly training both domain hint generation and domain translation tasks. Our experimental results demonstrate that our approach significantly outperforms direct fine-tuning in both in-domain and OOD scenarios.
\section*{Limitations}
The construction of a multi-domain translation test dataset presents challenges due to the limited availability of open-source data, which makes it difficult to ensure that the dataset is free from data leakage relative to LLMs. For example, LLaMA-2-70b demonstrates outstanding performance in the Bible domain. Moreover, although many domains within these datasets have been utilized in previous papers, the quality of the domains cannot be guaranteed. Secondly, our proposed method partly depends on the capabilities of the base model. In cases where the base model lacks a certain level of domain-specific knowledge, it may not be possible to extract useful information for translation, even with the Chain-of-Thought strategy.

\section*{Ethics Statement}
With the exception of the 9 domain test sets in Chinese$\Leftrightarrow$English sourced from internal data, all data and models utilized in this paper are open-source. And the models involved are all trained on public datasets. Our method and strategy do not involve ethical issues.

\section*{Acknowledgements}
The work was supported by Alibaba Innovative Research Program, the General Program of National Natural Science Foundation of China (62176153), Shanghai Municipal Science and Technology Major Project (2021SHZDZX0102), SMP-Zhipu.AI Large Model Cross-Disciplinary Fund, the Science and Technology Development Fund, Macau SAR (Grant Nos. FDCT/060/2022/AFJ, FDCT/0070/2022/AMJ), and the Multi-year Research Grant from the University of Macau (Grant No. MYRG-GRG2023-00006-FST-UMDF).

\normalem
\bibliography{anthology,acl2023}

\begin{thebibliography}{40}
\expandafter\ifx\csname natexlab\endcsname\relax\def\natexlab#1{#1}\fi

\bibitem[{Achiam et~al.(2023)Achiam, Adler, Agarwal, Ahmad, Akkaya, Aleman et~al.}]{Achiam2023GPT4TR}
OpenAI~Josh Achiam, Steven Adler, Sandhini Agarwal, Lama Ahmad, Ilge Akkaya, Florencia~Leoni Aleman, et~al. 2023.
\newblock \href {https://api.semanticscholar.org/CorpusID:257532815} {Gpt-4 technical report}.

\bibitem[{Agrawal et~al.(2023)Agrawal, Zhou, Lewis, Zettlemoyer, and Ghazvininejad}]{Agrawal2022IncontextES}
Sweta Agrawal, Chunting Zhou, Mike Lewis, Luke Zettlemoyer, and Marjan Ghazvininejad. 2023.
\newblock \href {https://doi.org/10.18653/v1/2023.findings-acl.564} {In-context examples selection for machine translation}.
\newblock In \emph{Findings of the Association for Computational Linguistics: ACL 2023}, pages 8857--8873, Toronto, Canada. Association for Computational Linguistics.

\bibitem[{Aharoni and Goldberg(2020)}]{Aharoni2020UnsupervisedDC}
Roee Aharoni and Yoav Goldberg. 2020.
\newblock \href {https://doi.org/10.18653/v1/2020.acl-main.692} {Unsupervised domain clusters in pretrained language models}.
\newblock In \emph{Proceedings of the 58th Annual Meeting of the Association for Computational Linguistics}, pages 7747--7763, Online. Association for Computational Linguistics.

\bibitem[{Alves et~al.(2023)Alves, Guerreiro, Alves, Pombal, Rei, de~Souza, Colombo, and Martins}]{Alves2023SteeringLL}
Duarte~M. Alves, Nuno~M. Guerreiro, Joao Alves, Jos{\'e}~P. Pombal, Ricardo Rei, Jos'e G.~C. de~Souza, Pierre Colombo, and Andr{\'e} Martins. 2023.
\newblock \href {https://api.semanticscholar.org/CorpusID:264405904} {Steering large language models for machine translation with finetuning and in-context learning}.
\newblock In \emph{Conference on Empirical Methods in Natural Language Processing}.

\bibitem[{Cettolo et~al.(2017)Cettolo, Federico, Bentivogli, Niehues, St{\"u}ker, Sudoh, Yoshino, and Federmann}]{cettolo-etal-2017-overview}
Mauro Cettolo, Marcello Federico, Luisa Bentivogli, Jan Niehues, Sebastian St{\"u}ker, Katsuhito Sudoh, Koichiro Yoshino, and Christian Federmann. 2017.
\newblock \href {https://aclanthology.org/2017.iwslt-1.1} {Overview of the {IWSLT} 2017 evaluation campaign}.
\newblock In \emph{Proceedings of the 14th International Conference on Spoken Language Translation}, pages 2--14, Tokyo, Japan. International Workshop on Spoken Language Translation.

\bibitem[{Cettolo et~al.(2014)Cettolo, Niehues, St{\"u}ker, Bentivogli, and Federico}]{cettolo-etal-2014-report}
Mauro Cettolo, Jan Niehues, Sebastian St{\"u}ker, Luisa Bentivogli, and Marcello Federico. 2014.
\newblock \href {https://aclanthology.org/2014.iwslt-evaluation.1} {Report on the 11th {IWSLT} evaluation campaign}.
\newblock In \emph{Proceedings of the 11th International Workshop on Spoken Language Translation: Evaluation Campaign}, pages 2--17, Lake Tahoe, California.

\bibitem[{Chu and Wang(2020)}]{Chu2020}
Chenhui Chu and Rui Wang. 2020.
\newblock \href {https://doi.org/10.2197/ipsjjip.28.413} {A survey of domain adaptation for machine translation}.
\newblock \emph{Journal of Information Processing}, 28:413--426.

\bibitem[{Currey et~al.(2020)Currey, Mathur, and Dinu}]{currey-etal-2020-distilling}
Anna Currey, Prashant Mathur, and Georgiana Dinu. 2020.
\newblock \href {https://doi.org/10.18653/v1/2020.emnlp-main.364} {Distilling multiple domains for neural machine translation}.
\newblock In \emph{Proceedings of the 2020 Conference on Empirical Methods in Natural Language Processing (EMNLP)}, pages 4500--4511, Online. Association for Computational Linguistics.

\bibitem[{French(1999)}]{French1999CatastrophicFI}
Robert~M. French. 1999.
\newblock \href {https://api.semanticscholar.org/CorpusID:2691726} {Catastrophic forgetting in connectionist networks}.
\newblock \emph{Trends in Cognitive Sciences}, 3:128--135.

\bibitem[{Ghazvininejad et~al.(2023)Ghazvininejad, Gonen, and Zettlemoyer}]{Ghazvininejad2023DictionarybasedPP}
Marjan Ghazvininejad, Hila Gonen, and Luke Zettlemoyer. 2023.
\newblock \href {https://arxiv.org/abs/2302.07856} {Dictionary-based phrase-level prompting of large language models for machine translation}.
\newblock \emph{ArXiv preprint}, abs/2302.07856.

\bibitem[{He et~al.(2023)He, Liang, Jiao, Zhang, Yang, Wang, Tu, Shi, and Wang}]{He2023ExploringHT}
Zhiwei He, Tian Liang, Wenxiang Jiao, Zhuosheng Zhang, Yujiu Yang, Rui Wang, Zhaopeng Tu, Shuming Shi, and Xing Wang. 2023.
\newblock \href {https://arxiv.org/abs/2305.04118} {Exploring human-like translation strategy with large language models}.
\newblock \emph{ArXiv preprint}, abs/2305.04118.

\bibitem[{Hu et~al.(2022)Hu, Shen, Wallis, Allen{-}Zhu, Li, Wang, Wang, and Chen}]{Hu2021LoRALA}
Edward~J. Hu, Yelong Shen, Phillip Wallis, Zeyuan Allen{-}Zhu, Yuanzhi Li, Shean Wang, Lu~Wang, and Weizhu Chen. 2022.
\newblock \href {https://openreview.net/forum?id=nZeVKeeFYf9} {Lora: Low-rank adaptation of large language models}.
\newblock In \emph{The Tenth International Conference on Learning Representations, {ICLR} 2022, Virtual Event, April 25-29, 2022}. OpenReview.net.

\bibitem[{Jiao et~al.(2023)Jiao, Huang, Wang, He, Liang, Wang, Shi, and Tu}]{Jiao2023ParroTTD}
Wenxiang Jiao, Jen-tse Huang, Wenxuan Wang, Zhiwei He, Tian Liang, Xing Wang, Shuming Shi, and Zhaopeng Tu. 2023.
\newblock \href {https://doi.org/10.18653/v1/2023.findings-emnlp.1001} {{P}arro{T}: Translating during chat using large language models tuned with human translation and feedback}.
\newblock In \emph{Findings of the Association for Computational Linguistics: EMNLP 2023}, pages 15009--15020, Singapore. Association for Computational Linguistics.

\bibitem[{Koehn and Knowles(2017)}]{koehn2017six}
Philipp Koehn and Rebecca Knowles. 2017.
\newblock \href {https://doi.org/10.18653/v1/W17-3204} {Six challenges for neural machine translation}.
\newblock In \emph{Proceedings of the First Workshop on Neural Machine Translation}, pages 28--39, Vancouver. Association for Computational Linguistics.

\bibitem[{Koneru et~al.(2023)Koneru, Exel, Huck, and Niehues}]{Koneru2023ContextualRO}
Sai Koneru, Miriam Exel, Matthias Huck, and Jan Niehues. 2023.
\newblock \href {https://arxiv.org/abs/2310.14855} {Contextual refinement of translations: Large language models for sentence and document-level post-editing}.
\newblock \emph{ArXiv preprint}, abs/2310.14855.

\bibitem[{Lai et~al.(2022)Lai, Libovick{\'y}, and Fraser}]{Lai2021ImprovingBD}
Wen Lai, Jind{\v{r}}ich Libovick{\'y}, and Alexander Fraser. 2022.
\newblock \href {https://aclanthology.org/2022.coling-1.461} {Improving both domain robustness and domain adaptability in machine translation}.
\newblock In \emph{Proceedings of the 29th International Conference on Computational Linguistics}, pages 5191--5204, Gyeongju, Republic of Korea. International Committee on Computational Linguistics.

\bibitem[{Lee et~al.(2022)Lee, Kim, Cho, Choi, and Park}]{lee-etal-2022-specializing}
Jiyoung Lee, Hantae Kim, Hyunchang Cho, Edward Choi, and Cheonbok Park. 2022.
\newblock \href {https://doi.org/10.18653/v1/2022.emnlp-main.680} {Specializing multi-domain {NMT} via penalizing low mutual information}.
\newblock In \emph{Proceedings of the 2022 Conference on Empirical Methods in Natural Language Processing}, pages 10015--10026, Abu Dhabi, United Arab Emirates. Association for Computational Linguistics.

\bibitem[{Li et~al.(2022)Li, Zhang, and Zhao}]{li2022self}
Junlong Li, Zhuosheng Zhang, and Hai Zhao. 2022.
\newblock \href {https://arxiv.org/abs/2212.08635} {Self-prompting large language models for open-domain qa}.
\newblock \emph{ArXiv preprint}, abs/2212.08635.

\bibitem[{Man et~al.(2023)Man, Zhang, Chen, Chen, and Xu}]{Man2023ExploringDA}
Zhibo Man, Yujie Zhang, Yuanmeng Chen, Yufeng Chen, and Jinan Xu. 2023.
\newblock \href {https://api.semanticscholar.org/CorpusID:265158343} {Exploring domain-shared and domain-specific knowledge in multi-domain neural machine translation}.
\newblock In \emph{Machine Translation Summit}.

\bibitem[{Moslem et~al.(2023)Moslem, Haque, Kelleher, and Way}]{Moslem2023AdaptiveMT}
Yasmin Moslem, Rejwanul Haque, John~D. Kelleher, and Andy Way. 2023.
\newblock \href {https://aclanthology.org/2023.eamt-1.22} {Adaptive machine translation with large language models}.
\newblock In \emph{Proceedings of the 24th Annual Conference of the European Association for Machine Translation}, pages 227--237, Tampere, Finland. European Association for Machine Translation.

\bibitem[{Ouyang et~al.(2022)Ouyang, Wu, Jiang, Almeida, Wainwright, Mishkin, Zhang, Agarwal, Slama, Ray, Schulman, Hilton, Kelton, Miller, Simens, Askell, Welinder, Christiano, Leike, and Lowe}]{Ouyang2022TrainingLM}
Long Ouyang, Jeffrey Wu, Xu~Jiang, Diogo Almeida, Carroll Wainwright, Pamela Mishkin, Chong Zhang, Sandhini Agarwal, Katarina Slama, Alex Ray, John Schulman, Jacob Hilton, Fraser Kelton, Luke Miller, Maddie Simens, Amanda Askell, Peter Welinder, Paul~F Christiano, Jan Leike, and Ryan Lowe. 2022.
\newblock \href {https://proceedings.neurips.cc/paper_files/paper/2022/file/b1efde53be364a73914f58805a001731-Paper-Conference.pdf} {Training language models to follow instructions with human feedback}.
\newblock In \emph{Advances in Neural Information Processing Systems}, volume~35, pages 27730--27744. Curran Associates, Inc.

\bibitem[{Papineni et~al.(2002)Papineni, Roukos, Ward, and Zhu}]{Papineni2002BleuAM}
Kishore Papineni, Salim Roukos, Todd Ward, and Wei-Jing Zhu. 2002.
\newblock \href {https://doi.org/10.3115/1073083.1073135} {{B}leu: a method for automatic evaluation of machine translation}.
\newblock In \emph{Proceedings of the 40th Annual Meeting of the Association for Computational Linguistics}, pages 311--318, Philadelphia, Pennsylvania, USA. Association for Computational Linguistics.

\bibitem[{Pham et~al.(2021)Pham, Crego, and Yvon}]{Pham2021}
MinhQuang Pham, Josep~Maria Crego, and Fran{\c{c}}ois Yvon. 2021.
\newblock \href {https://doi.org/10.1162/tacl_a_00351} {Revisiting multi-domain machine translation}.
\newblock \emph{Transactions of the Association for Computational Linguistics}, 9:17--35.

\bibitem[{Post(2018)}]{Post2018ACF}
Matt Post. 2018.
\newblock \href {https://doi.org/10.18653/v1/W18-6319} {A call for clarity in reporting {BLEU} scores}.
\newblock In \emph{Proceedings of the Third Conference on Machine Translation: Research Papers}, pages 186--191, Brussels, Belgium. Association for Computational Linguistics.

\bibitem[{Raunak et~al.(2023)Raunak, Sharaf, Wang, Awadalla, and Menezes}]{Raunak2023LeveragingGF}
Vikas Raunak, Amr Sharaf, Yiren Wang, Hany Awadalla, and Arul Menezes. 2023.
\newblock \href {https://doi.org/10.18653/v1/2023.findings-emnlp.804} {Leveraging {GPT}-4 for automatic translation post-editing}.
\newblock In \emph{Findings of the Association for Computational Linguistics: EMNLP 2023}, pages 12009--12024, Singapore. Association for Computational Linguistics.

\bibitem[{Rei et~al.(2020)Rei, Stewart, Farinha, and Lavie}]{Rei2020COMETAN}
Ricardo Rei, Craig Stewart, Ana~C Farinha, and Alon Lavie. 2020.
\newblock \href {https://doi.org/10.18653/v1/2020.emnlp-main.213} {{COMET}: A neural framework for {MT} evaluation}.
\newblock In \emph{Proceedings of the 2020 Conference on Empirical Methods in Natural Language Processing (EMNLP)}, pages 2685--2702, Online. Association for Computational Linguistics.

\bibitem[{Saunders(2021)}]{Saunders2021DomainAA}
Danielle Saunders. 2021.
\newblock \href {https://api.semanticscholar.org/CorpusID:233231665} {Domain adaptation and multi-domain adaptation for neural machine translation: A survey}.
\newblock \emph{J. Artif. Intell. Res.}, 75:351--424.

\bibitem[{Taori et~al.(2023)Taori, Gulrajani, Zhang, Dubois, Li, Guestrin, Liang, and Hashimoto}]{alpaca}
Rohan Taori, Ishaan Gulrajani, Tianyi Zhang, Yann Dubois, Xuechen Li, Carlos Guestrin, Percy Liang, and Tatsunori~B. Hashimoto. 2023.
\newblock Stanford alpaca: An instruction-following llama model.
\newblock \url{https://github.com/tatsu-lab/stanford_alpaca}.

\bibitem[{Thompson et~al.(2019)Thompson, Gwinnup, Khayrallah, Duh, and Koehn}]{Thompson2019OvercomingCF}
Brian Thompson, Jeremy Gwinnup, Huda Khayrallah, Kevin Duh, and Philipp Koehn. 2019.
\newblock \href {https://doi.org/10.18653/v1/N19-1209} {Overcoming catastrophic forgetting during domain adaptation of neural machine translation}.
\newblock In \emph{Proceedings of the 2019 Conference of the North {A}merican Chapter of the Association for Computational Linguistics: Human Language Technologies, Volume 1 (Long and Short Papers)}, pages 2062--2068, Minneapolis, Minnesota. Association for Computational Linguistics.

\bibitem[{Tian et~al.(2014)Tian, Wong, Chao, Quaresma, Oliveira, Lu, Li, Wang, and Wang}]{Tian2014UMCorpusAL}
Liang Tian, Derek~F. Wong, Lidia~S. Chao, Paulo Quaresma, Francisco Oliveira, Yi~Lu, Shuo Li, Yiming Wang, and Longyue Wang. 2014.
\newblock \href {http://www.lrec-conf.org/proceedings/lrec2014/pdf/774_Paper.pdf} {{UM}-corpus: A large {E}nglish-{C}hinese parallel corpus for statistical machine translation}.
\newblock In \emph{Proceedings of the Ninth International Conference on Language Resources and Evaluation ({LREC}'14)}, pages 1837--1842, Reykjavik, Iceland. European Language Resources Association (ELRA).

\bibitem[{Vaswani et~al.(2017)Vaswani, Shazeer, Parmar, Uszkoreit, Jones, Gomez, Kaiser, and Polosukhin}]{NIPS2017_3f5ee243}
Ashish Vaswani, Noam Shazeer, Niki Parmar, Jakob Uszkoreit, Llion Jones, Aidan~N. Gomez, Lukasz Kaiser, and Illia Polosukhin. 2017.
\newblock \href {https://proceedings.neurips.cc/paper/2017/hash/3f5ee243547dee91fbd053c1c4a845aa-Abstract.html} {Attention is all you need}.
\newblock In \emph{Advances in Neural Information Processing Systems 30: Annual Conference on Neural Information Processing Systems 2017, December 4-9, 2017, Long Beach, CA, {USA}}, pages 5998--6008.

\bibitem[{Wang et~al.(2022)Wang, Kordi, Mishra, Liu, Smith, Khashabi, and Hajishirzi}]{Wang2022SelfInstructAL}
Yizhong Wang, Yeganeh Kordi, Swaroop Mishra, Alisa Liu, Noah~A. Smith, Daniel Khashabi, and Hannaneh Hajishirzi. 2022.
\newblock \href {https://api.semanticscholar.org/CorpusID:254877310} {Self-instruct: Aligning language models with self-generated instructions}.
\newblock In \emph{Annual Meeting of the Association for Computational Linguistics}.

\bibitem[{Wang et~al.(2019)Wang, Wang, Shi, Li, and Tu}]{Wang2019GoFT}
Yong Wang, Longyue Wang, Shuming Shi, Victor O.~K. Li, and Zhaopeng Tu. 2019.
\newblock \href {https://api.semanticscholar.org/CorpusID:208248157} {Go from the general to the particular: Multi-domain translation with domain transformation networks}.
\newblock In \emph{AAAI Conference on Artificial Intelligence}.

\bibitem[{Wei et~al.(2022)Wei, Wang, Schuurmans, Bosma, hsin Chi, Xia, Le, and Zhou}]{Wei2022ChainOT}
Jason Wei, Xuezhi Wang, Dale Schuurmans, Maarten Bosma, Ed~Huai hsin Chi, F.~Xia, Quoc Le, and Denny Zhou. 2022.
\newblock \href {http://papers.nips.cc/paper\_files/paper/2022/hash/9d5609613524ecf4f15af0f7b31abca4-Abstract-Conference.html} {Chain of thought prompting elicits reasoning in large language models}.
\newblock In \emph{Advances in Neural Information Processing Systems 35: Annual Conference on Neural Information Processing Systems 2022, NeurIPS 2022, New Orleans, LA, USA, November 28 - December 9, 2022}.

\bibitem[{Xu et~al.(2023)Xu, Kim, Sharaf, and Awadalla}]{Xu2023APS}
Haoran Xu, Young~Jin Kim, Amr Sharaf, and Hany~Hassan Awadalla. 2023.
\newblock \href {https://arxiv.org/abs/2309.11674} {A paradigm shift in machine translation: Boosting translation performance of large language models}.
\newblock \emph{ArXiv preprint}, abs/2309.11674.

\bibitem[{Zeng et~al.(2023)Zeng, Meng, Yin, and Zhou}]{Zeng2023TIMTL}
Jiali Zeng, Fandong Meng, Yongjing Yin, and Jie Zhou. 2023.
\newblock \href {https://arxiv.org/abs/2307.04408} {Tim: Teaching large language models to translate with comparison}.
\newblock \emph{ArXiv preprint}, abs/2307.04408.

\bibitem[{Zhan et~al.(2021)Zhan, Liu, Wong, and Chao}]{zhan2021metacl}
Runzhe Zhan, Xuebo Liu, Derek~F. Wong, and Lidia~S. Chao. 2021.
\newblock \href {https://ojs.aaai.org/index.php/AAAI/article/view/17683} {Meta-curriculum learning for domain adaptation in neural machine translation}.
\newblock In \emph{Thirty-Fifth {AAAI} Conference on Artificial Intelligence, {AAAI} 2021, Thirty-Third Conference on Innovative Applications of Artificial Intelligence, {IAAI} 2021, The Eleventh Symposium on Educational Advances in Artificial Intelligence, {EAAI} 2021, Virtual Event, February 2-9, 2021}, pages 14310--14318. {AAAI} Press.

\bibitem[{Zhang et~al.(2023{\natexlab{a}})Zhang, Haddow, and Birch}]{Zhang2023PromptingLL}
Biao Zhang, Barry Haddow, and Alexandra Birch. 2023{\natexlab{a}}.
\newblock \href {https://proceedings.mlr.press/v202/zhang23m.html} {Prompting large language model for machine translation: A case study}.
\newblock In \emph{Proceedings of the 40th International Conference on Machine Learning}, volume 202 of \emph{Proceedings of Machine Learning Research}, pages 41092--41110. PMLR.

\bibitem[{Zhang et~al.(2022)Zhang, Yang, Wei, Liu, Fan, Si, and Xie}]{zhang-etal-2022-competency}
Pei Zhang, Baosong Yang, Hao-Ran Wei, Dayiheng Liu, Kai Fan, Luo Si, and Jun Xie. 2022.
\newblock \href {https://doi.org/10.18653/v1/2022.emnlp-main.330} {Competency-aware neural machine translation: Can machine translation know its own translation quality?}
\newblock In \emph{Proceedings of the 2022 Conference on Empirical Methods in Natural Language Processing}. Association for Computational Linguistics.

\bibitem[{Zhang et~al.(2023{\natexlab{b}})Zhang, Fang, Zhang, Ma, Zhou, Huang, Bu, Gui, Chen, Chen, and Feng}]{bayling}
Shaolei Zhang, Qingkai Fang, Zhuocheng Zhang, Zhengrui Ma, Yan Zhou, Langlin Huang, Mengyu Bu, Shangtong Gui, Yunji Chen, Xilin Chen, and Yang Feng. 2023{\natexlab{b}}.
\newblock \href {https://arxiv.org/abs/2306.10968} {Bayling: Bridging cross-lingual alignment and instruction following through interactive translation for large language models}.
\newblock \emph{ArXiv preprint}, abs/2306.10968.

\end{thebibliography}
\bibliographystyle{acl_natbib}

\appendix

\section{Details of Multi-domain Test Sets}
\label{sec:appendix_data}
The detailed data description and source are as follows:
\begin{itemize}  
    \item De$\Leftrightarrow$En: We collect a total of 25 test sets from various sources, including 5 test sets from~\citet{Aharoni2020UnsupervisedDC} (Medical, Law, IT, Koran and Subtitles), 9 test sets from the OPUS website\footnote{\url{https://opus.nlpl.eu}} (Bible, Book, ECB, Global Voices, QED, RF, MultiUN, TED 2020, Tilde), 1 test set from~\citet{zhan2021metacl} \footnote{\url{https://github.com/NLP2CT/Meta-Curriculum}} (Covid-19 News), 2 test sets from IWSLT official competitions~\citep{cettolo-etal-2014-report, cettolo-etal-2017-overview} (Tedtalk14 and Tedtalk17), 8 test sets from WMT official competitions\footnote{\label{wmt}\url{https://www.statmt.org}}, including News(WMT14) from WMT14 Machine Translation, Mixed(WMT22) from WMT22 General Machine Translation, Chat(WMT20) from WMT20 Machine Translation for Chats, Chat(WMT22) from WMT22 Chat Shared Task, Medical(WMT14) from WMT14 Medical Translation, Bio(WMT20) from WMT20 Biomedical Translation Task, Bio(WMT22) from WMT22 Biomedical Translation Task, IT(WMT16) from WMT16 Machine Translation of IT domain. For the test sets obtained from OPUS, we randomly select 2000 samples from the related domain dataset, as there were no pre-defined test sets. 
    \item Zh$\Leftrightarrow$En: We collect a total of 22 test sets from various sources, including 4 test sets from UM-Corpus~\citep{Tian2014UMCorpusAL} (News, Science, Laws, Subtitles), 1 test sets from IWSLT17~\citep{cettolo-etal-2017-overview} official competitions (Tedtalk17), 7 test sets from internal test set (Medicine, IT, ITELEC, Finance, Car, Machine, Energyminera), 10 test sets from WMT official competitions, including Mix(WMT22) from WMT22 General Machine Translation (and four sub-areas for its subdivision: News(WMT22), Social(WMT22), Ecommerce(WMT22), and Conversational(WMT22)), News(WMT19) from WMT19 Machine Translation of News, Bio(WMT20) from WMT20 Biomedical Translation, Bio(WMT22) from WMT22 Biomedical Translation Task, Webnovel1 and Webnovel2 from WMT23 Discourse-Level Literary Translation test1 and test2.
\end{itemize}  

The size of each test set are in Table \ref{tab:data num de-en}, \ref{tab:data num en-de}, \ref{tab:data num zh-en} and \ref{tab:data num en-zh}. We used the entire dataset during testing.

\begin{table}[htbp]
\centering
\resizebox{0.45\textwidth}{!}{
\begin{tabular}{cc|cc|cc}
\toprule
Test set & Num & Test set & Num & Test set & Num \\
\hline
IT(OPUS) & 2000 & Medical & 2000 & Law & 2000 \\
Subtitles & 2000 & Koran & 2000 & Bible & 2000 \\
Covid & 3325 & ECB & 2000 & Globalvoices & 2000 \\
News(WMT14) & 3003 & Mixed(WMT22) & 1984 & Tedtalk14 & 1305 \\
Tedtalk17 & 1138 & TED2020 & 2000 & Chat(WMT20) & 967 \\
Chat(WMT22) & 1334 & Medical(WMT14) & 1000 & Bio(WMT20) & 404 \\
Bio(WMT22) & 358 & IT(WMT16) & 1000 & Book & 2000 \\
RF & 151 & Multiun & 2000 & Tilde & 2000 \\
QED & 2000 &  &  &  &   \\
\bottomrule
\end{tabular}
}
\caption{The size of each German$\Rightarrow$English test set.}
 \label{tab:data num de-en}
\end{table}

\begin{table}[htbp]
\centering
\resizebox{0.45\textwidth}{!}{
\begin{tabular}{cc|cc|cc}
\toprule
Test set & Num & Test set & Num & Test set & Num \\
\hline
IT(OPUS) & 2000 & Medical & 2000 & Law & 2000 \\
Subtitles & 2000 & Koran & 2000 & Bible & 2000 \\
Covid & 3325 & ECB & 2000 & Globalvoices & 2000 \\
News(WMT14) & 3003 & Mixed(WMT22) & 2037 & Tedtalk14 & 1305 \\
Tedtalk17 & 1138 & TED2020 & 2000 & Chat(WMT20) & 1133 \\
Chat(WMT22) & 1154 & Medical(WMT14) & 1000 & Bio(WMT20) & 505 \\
Bio(WMT22) & 383 & IT(WMT16) & 1000 & Book & 2000 \\
RF & 151 & Multiun & 2000 & Tilde & 2000 \\
QED & 2000 &  &  &  & \\
\bottomrule
\end{tabular}
}
\caption{The size of each English$\Rightarrow$German test set.}
 \label{tab:data num en-de}
\end{table}

\begin{table}[htbp]
\centering
\resizebox{0.45\textwidth}{!}{
\begin{tabular}{cc|cc|cc}
\toprule
Test set & Num & Test set & Num & Test set & Num \\
\hline
News & 1500 & Science & 503 & Law & 456 \\
Subtitles & 597 & Mixed(WMT22) & 1875 & News(WMT22) & 505 \\
Social(WMT22) & 503 & Conversation(WMT22) & 349 & Ecommerce(WMT22) & 518 \\
News(WMT19) & 2000 & Medicine & 485 & Tedtalk17 & 1459 \\
IT & 2011 & ITELEC & 1755 & Finance & 719 \\
Car & 469 & Machine & 1538 & Energyminera & 1480 \\
Webnovel1 & 645 & Webnovel2 & 869 & Bio(WMT20) & 300 \\
Bio(WMT22) & 264 &  &  &  & \\
\bottomrule
\end{tabular}
}
\caption{The size of each Chinese$\Rightarrow$English test set.}
 \label{tab:data num zh-en}
\end{table}

\begin{table}[htbp]
\centering
\resizebox{0.45\textwidth}{!}{
\begin{tabular}{cc|cc|cc}
\toprule
Test set & Num & Test set & Num & Test set & Num \\
\hline
News & 1500 & Science & 503 & Law & 456 \\
Subtitles & 597 & Mixed(WMT22) & 2037 & News(WMT22) & 511 \\
Social(WMT22) & 512 & Conversation(WMT22) & 484 & Ecommerce(WMT22) & 530 \\
News(WMT19) & 1997 & Medicine & 485 & Tedtalk17 & 1459 \\
IT & 2011 & ITELEC & 1755 & Finance & 719 \\
Car & 469 & Machine & 1538 & Energyminera & 1480 \\
Webnovel1 & 645 & Webnovel2 & 869 & Bio(WMT20) & 240 \\
Bio(WMT22) & 346 &  &  &  &  \\
\bottomrule
\end{tabular}
}
\caption{The size of each English$\Rightarrow$Chinese test set.}
 \label{tab:data num en-zh}
\end{table}

\section{Prompt Format for LLMs}
\label{sec:appendix_prompt}
This section presents the exact translation prompts for LLMs. We use the German-to-English translation as an example.

For ChatGPT, GPT4, LLaMA-2-70b and BM25, we use 1-shot prompt as following:

\begin{quote}
\textit{Translate the following sentence from German into English.}

\textit{German: Schaufensterpuppen, die dank Videoprojektionen erschreckend menschlich aussehen.}

\textit{English: Mannequins which look alarmingly human thanks to video projection.}

\textit{}

\textit{Translate the following sentence from German into English.}

\textit{German: Die Silberpfeile vermuten gleich zwei Regelbrüche im letzten Grand Prix des Jahres.}

\textit{English:}
\end{quote}

For BLOOMZ-7b1 and LLaMA-2-7b, we use the 5-shot prompt as follows:
\begin{quote}
\textit{Translate the following sentence from German into English.}

\textit{German: Schaufensterpuppen, die dank Videoprojektionen erschreckend menschlich aussehen.}

\textit{English: Mannequins which look alarmingly human thanks to video projection.}

\textit{}

\textit{...}

\textit{}

\textit{Translate the following sentence from German into English.}

\textit{German: Die Silberpfeile vermuten gleich zwei Regelbrüche im letzten Grand Prix des Jahres.}

\textit{English:}
\end{quote}

For ParroT-7b, we use the prompt provided by the default settings of the model with no error hint.

\begin{quote}
\textit{\#\#\# Instruction:}

\textit{Translate the following sentences from German to English.}

\textit{}

\textit{\#\#\# Input:}

\textit{Die Silberpfeile vermuten gleich zwei Regelbrüche im letzten Grand Prix des Jahres.}

\textit{}

\textit{\#\#\# Hint: A translation with no errors could be}

\textit{}

\textit{\#\#\# Response:}
\end{quote}

For ALMA-7b, we use the prompt provided by \citet{Xu2023APS}:

\begin{quote}
\textit{Translate this from German to English:}

\textit{German: Die Silberpfeile vermuten gleich zwei Regelbrüche im letzten Grand Prix des Jahres.}

\textit{English:}
\end{quote}

For BayLing-7b, we use the prompt as follows:

\begin{quote}
\textit{Translate the following sentence from German into English.}

\textit{German: Die Silberpfeile vermuten gleich zwei Regelbrüche im letzten Grand Prix des Jahres.}

\textit{English:}
\end{quote}

For MAPS, we followed the prompt provided by ~\citet{He2023ExploringHT}. To achieve better results and to ensure a fair comparison with other models, we utilized the 5-shot translation results from LLaMA-2-7b as the prompt for the candidate base in the MAPS method. 

For LLaMA-2-7b + FT + MAPS, the candidate base utilized the translation results from LLaMA-2-7b + FT.

For LLaMA-2-7b + FT, we use the translation prompt as follows:
\begin{quote}
\textit{\#\#\# Instruction:}

\textit{Translate the following German text into English.}

\textit{}

\textit{\#\#\# Input:}

\textit{Die Silberpfeile vermuten gleich zwei Regelbrüche im letzten Grand Prix des Jahres.}

\textit{}

\textit{\#\#\# Response:}
\end{quote}

\section{Details of Fine-tuning LLMs}
\label{sec:appendix_sft}

\begin{figure}[h]
  \centering
  \includegraphics[width=0.47\textwidth]{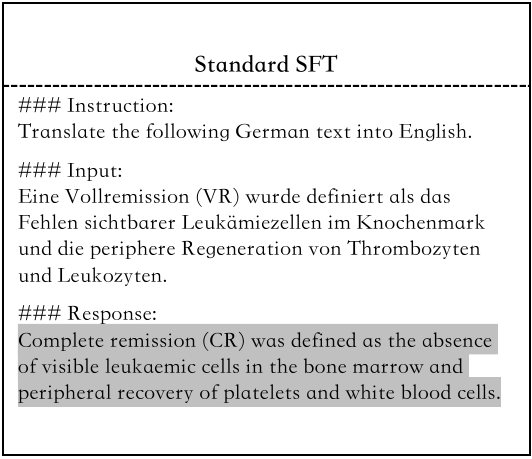}
  \caption{The detailed prompt format of fine-tuning. The gray areas are the parts that require training.}
  \label{fig:prompt_SFT}
\end{figure}
\begin{figure}[H]
  \centering
  \includegraphics[width=0.47\textwidth]{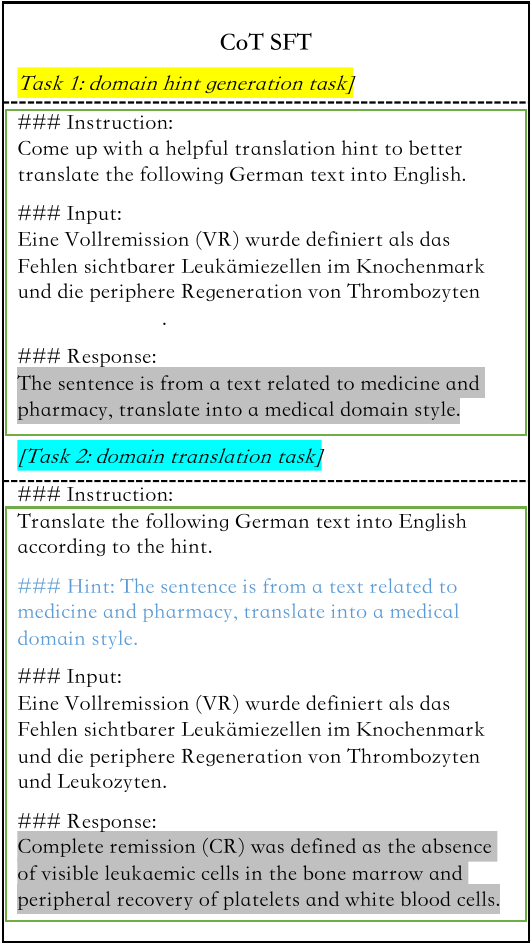}
  \caption{The detailed prompt format of CoT-FT splits into two tasks. The gray areas are what should be trained, and the blue font provides a domain hint.}
  \label{fig:prompt_CoT_SFT}
\end{figure}

Figure \ref{fig:prompt_SFT} shows the detailed prompt format for direct fine-tuning. The gray areas are the parts that require training.

The detailed prompt format of our proposed CoT-FT is shown in Figure \ref{fig:prompt_SFT}. The data consists of two parts: the first part is the domain hint prediction and generation task, and the second part is the translation task based on the generated domain hint.  

In the domain generation task, for German$\Leftrightarrow$English, the 10 monolingual domains we use are "IT(OPUS)", "Medical(OPUS)", "Law", "Subtitles", "Koran", "Bible", "ECB", "Global Voices", "Tedtalk14", and "Covid". For Chinese$\Leftrightarrow$English, the 8 monolingual domains we use are "News", "Science", "Law", "Subtitles", "Twitter", "Medical", "IT", "News(WMT19)". We manually write a domain-related hint based on the characteristics of each domain.

\paragraph{The details of manually crafted hints}
A translation domain hint refers to specific insights or contextual cues provided to translators or translation software to enhance the accuracy and appropriateness of translations within a particular field or subject matter. These hints can include vocabulary preferences, stylistic considerations unique to the domain, and cultural nuances that are pivotal in ensuring the translated content is both accurate and resonate with the target audience. Translation domain hints are crucial in specialized fields such as legal, medical, technical, and scientific translations, where precise terminology and context are paramount. In our experiments, the domain hint mainly describes the specific domain of the sentence and the stylistic characteristics of this domain. 

The manually crafted hints mainly describe the domain style characteristics. It follows the format as follows: \textit{"The sentence is from/about XXX. The style of text is XXX. Translate into a XXX domain style."}. \textit{XXX} is manually crafted by us according to the stylistic features of the current domain. For example: \textit{"The sentence is from the informal news stories. The style of text is informal, humanistic ... It is generally written in the first or second person ... Translate into this kind of news domain style."} Our intention is to offer a general approach for CoT fine-tuning; the style of the human-crafted hints can be determined according to personal needs and judgment.

Although there are only ten kinds of domain hints, the domains often have correlations among them. For instance, pharmacology, biology, and medicine are interrelated. Thus, a medical translation hint can also be beneficial for biology translation. Therefore, learning these ten domains' hint information can enhance translation performance in more than just these ten areas.

\paragraph{Case study}

As shown in Table \ref{tab:case_study}, there are some cases of the manual hint and the automatically generated hint for comparison. We can observe that for in-domain sentences, a hint identical to the manual hint is generated. However, for out-of-domain (OOD) sentences, a hint that is different from but still related to the manual hint is produced.

\section{Performance of Different Models on The Multi-domain Test Sets}
\label{sec:appendix_results}
\begin{figure*}[bpht]
\begin{minipage}[t]{0.5\textwidth}
\centering
\includegraphics[width=1\textwidth]{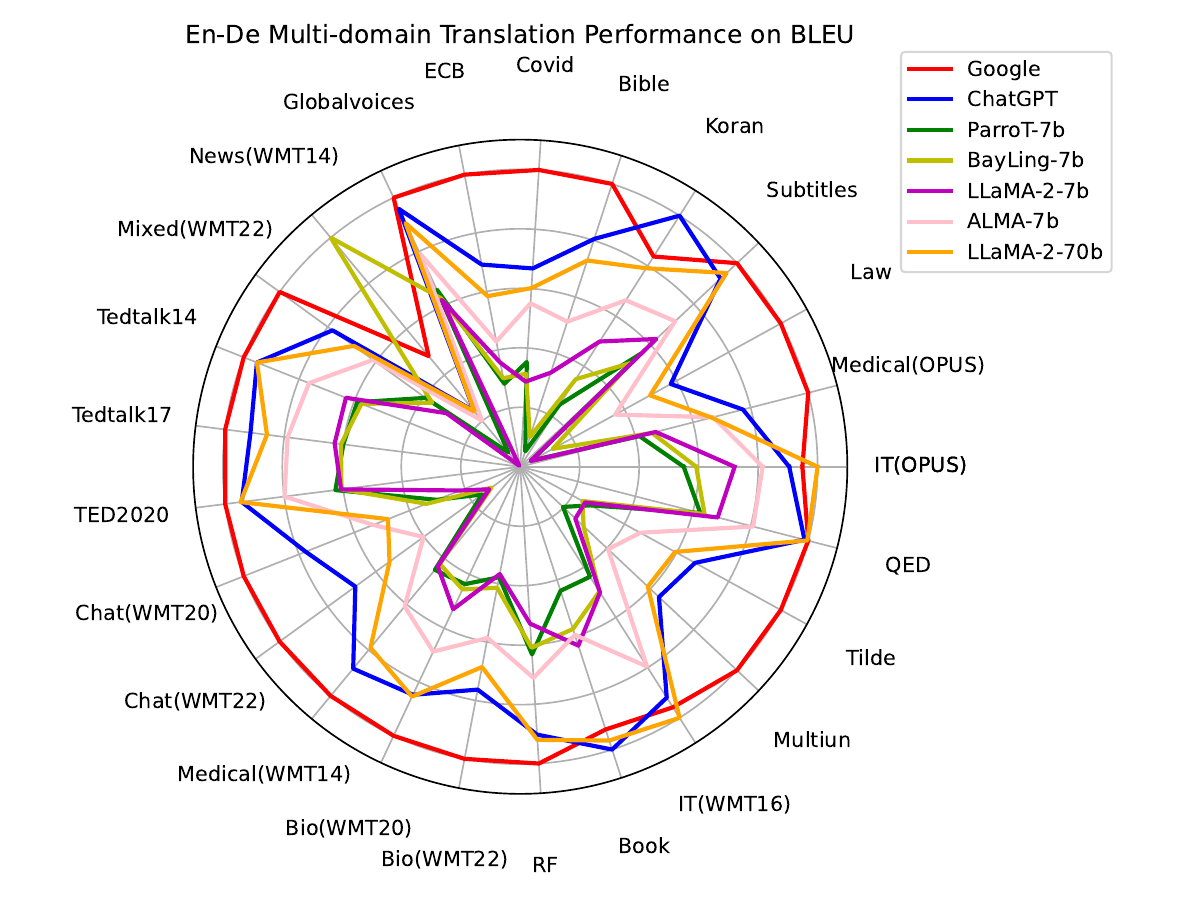}
\end{minipage}
\hfill
\begin{minipage}[t]{0.5\textwidth}
\centering
\includegraphics[width=1\textwidth]{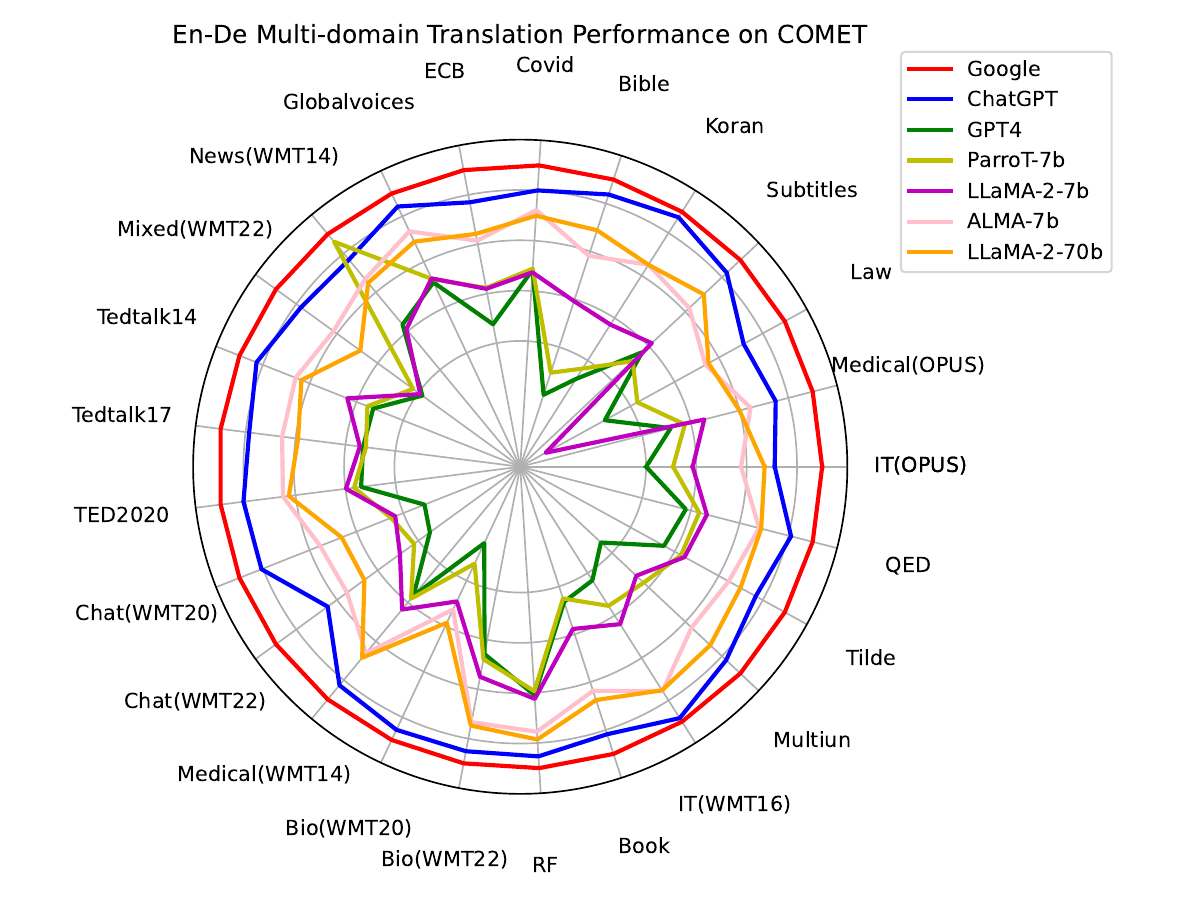}
\end{minipage}
 \caption{Performance comparison of prominent LLMs on the multi-domain English-to-German translation. For a clear comparison, we show the scores normalized by the maximum score in each domain. The performance of LLMs varies greatly across multi-domains. Best reviewed in colors.}

\label{fig:radar_en_de}
\end{figure*}

\begin{figure*}[bpht]
\begin{minipage}[t]{0.5\textwidth}
\centering
\includegraphics[width=1\textwidth]{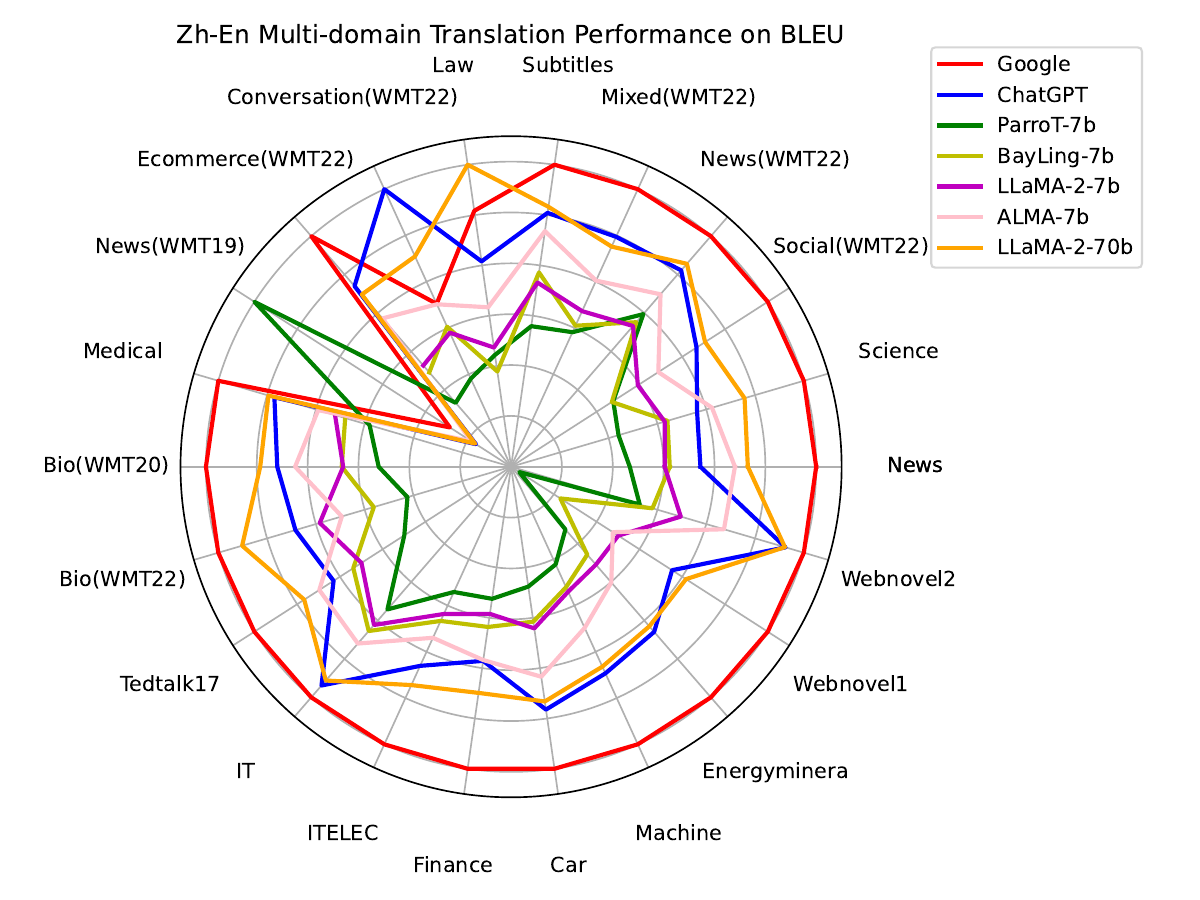}
\end{minipage}
\hfill
\begin{minipage}[t]{0.5\textwidth}
\centering
\includegraphics[width=1\textwidth]{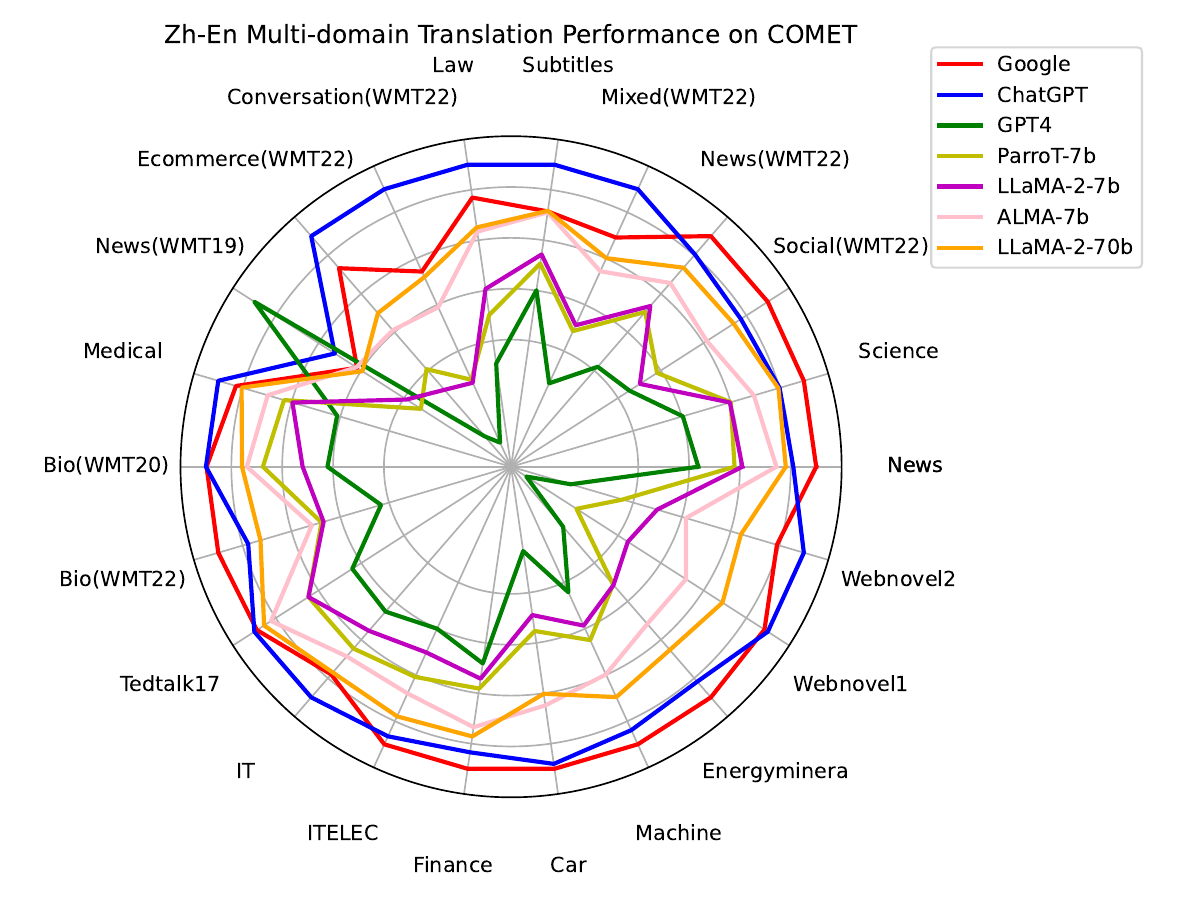}
\end{minipage}
 \caption{Performance comparision of prominent LLMs on the multi-domain Chinese-to-English translation. For clear comparison, we show the scores normalized by the maximum score in each domain. The performance of LLMs varies greatly across multi-domains. Best reviewed in colors.}

\label{fig:radar_zh_en}
\end{figure*}

\begin{figure*}[bpht]
\begin{minipage}[t]{0.5\textwidth}
\centering
\includegraphics[width=1\textwidth]{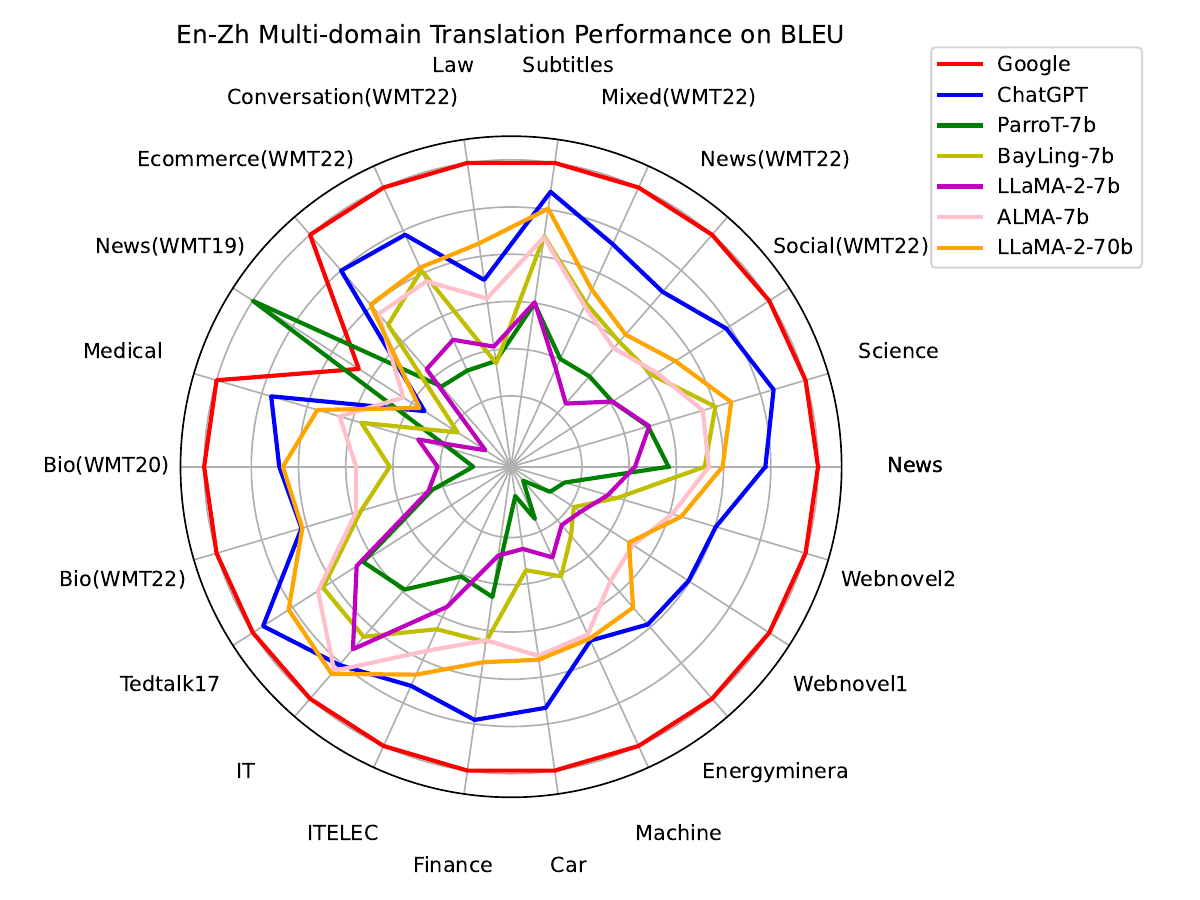}
\end{minipage}
\hfill
\begin{minipage}[t]{0.5\textwidth}
\centering
\includegraphics[width=1\textwidth]{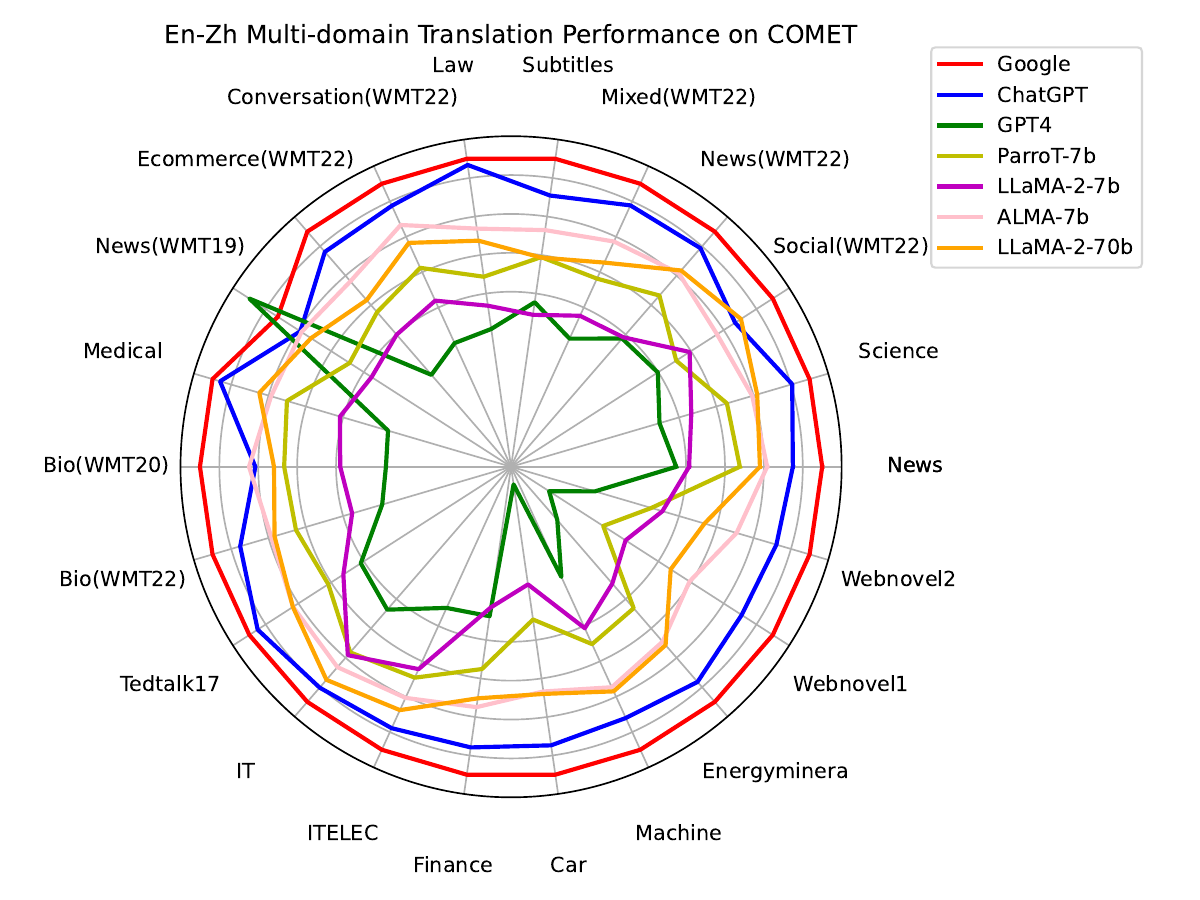}
\end{minipage}
 \caption{Performance comparision of prominent LLMs on the multi-domain English-to-Chinese translation. For clear comparison, we show the scores normalized by the maximum score in each domain. The performance of LLMs varies greatly across multi-domains. Best reviewed in colors.}

\label{fig:radar_en_zh}
\end{figure*}

Detailed results of all models on multi-domain translation test sets with BLEU and COMET are shown in Table \ref{tab:details_de_en_bleu}, \ref{tab:details_de_en_comet}, \ref{tab:details_en_de_bleu}, \ref{tab:details_en_de_comet}, \ref{tab:details_zh_en_bleu}, \ref{tab:details_zh_en_comet}, \ref{tab:details_en_zh_bleu} and \ref{tab:details_en_zh_comet}. The '/' in the table indicates that the corresponding model may have data leakage in this test set, so no results have been released. Ang Figure \ref{fig:radar_en_de}, \ref{fig:radar_zh_en}, \ref{fig:radar_en_zh} show a visual comparison of different models in each domain.

\section{The performance of different methods on LLaMA-2-13b}

As shown in Figure \ref{tab:model size scaling 13b}, under the scaling of the LLaMA-2-13b setting, CoT-FT has achieved an advantage in both BLEU and COEMT metrics.

\begin{table}[htbp]
\centering
\resizebox{0.45\textwidth}{!}{
\begin{tabular}{l|cccc}
\toprule
\multirow{2}{*}{\textbf{Method}} & \multicolumn{2}{c}{In-domain}&\multicolumn{2}{c}{OOD}\\
&BLEU&COMET&BLEU&COMET\\
\hline
        LLaMA-2-13b & 34.70 & 81.55 & 35.45 & 84.64 \\ 
        + FT & 44.31 & 84.76 & 37.69 & 84.93 \\ 
        + FT-BM25 & 44.76 & 84.80 & 37.55 & 84.82 \\
        + FT-MAPS & 41.60 & 84.91 & 36.71 & \textbf{85.52} \\
        + CoT-FT & \textbf{44.31} & \textbf{84.95} & \textbf{38.36} & 85.31 \\ 
\bottomrule
\end{tabular}
}
\caption{Performance comparison of different methods at 7b, 13b, 70b model sizes (with a training data size of 4*100k). CoT fine-tuning significantly outperforms direct fine-tuning across different model sizes.}
 \label{tab:model size scaling 13b}
\end{table}

\onecolumn
    \begin{longtable}{lp{11cm}}
    \hline
    in domain text & 然而，在培养物和生物工程移植物中重建带有毛囊和腺体的皮肤仍是一项尚未解决的生物医学挑战。 \\
    manual hint & The sentence is from a text related to medicine, pharmacy, chemistry and biology experiments. Translate into a medical domain style. \\
    automatical hint & The sentence is from a text related to medicine, pharmacy, chemistry and biology experiments. Translate into a medical domain style. \\
    \hline
    OOD text & 帮助用户通过整合多种数据来源，存储用户行为数据，构建用户画像，实时存储在Cassandra中，提供大数据风控、推荐等服务。 \\
    manual hint & The sentence is from Cloud IT domain. It mainly involves computer-related software development and usage methods, including many terms related to computer software and hardware. Pay attention to professional troubleshooting terminologies and sentence patterns when translating. Translate into this IT domain style.\\
    automatical hint & The sentence is about computer network security, translate into this Internet Technology domain style. \\
    \hline
    OOD text & 因为这种元件具有未达居里点前电阻随温度变化非常缓慢，具有恒温、调温和自动控温的功能，只发热，不发红，无明火，不易燃烧。 \\
    manual hint & The sentence about electronic equipment and IT, including microelectronics technology, circuits, characteristics of electronic materials, etc. Pay attention to relevant electronic terms when translating. Translate into this electronic domain style. \\
    automatical hint & The sentence is from the texts consisting of parallel terminologies and sentences in science and technology areas. Translate into this science and technology domain style. \\
    
    \hline
    \caption{Case study for the manual hint and the automatically generated hint for comparison.} 
    \label{tab:case_study} \\
    \end{longtable}
\twocolumn

\begin{table*}[htbp]
\centering
\resizebox{\textwidth}{!}{
\begin{tabular}{l|ccccccccc}
\toprule
        ~ & Google & ChatGPT & GPT4 & ParroT-7b & BayLing-7b & LLaMA-2-7b & ALMA-7b & LLaMA-2-70b & LLaMA-2-7b + FT  \\ \hline
        IT(OPUS) & 40.40 & 36.05 & 36.40 & 32.76 & 33.69 & 34.78 & 35.33 & 37.95 & 40.18  \\ 
        Medical & 48.49 & 40.28 & 38.75 & 37.77 & 38.60 & 38.11 & 39.59 & 40.14 & 45.46  \\
        Law & 46.58 & 37.80 & 35.82 & 30.86 & 33.40 & 33.67 & 36.40 & 43.10 & 46.12  \\
        Subtitles & 31.03 & 29.21 & 29.97 & 26.40 & 26.81 & 26.64 & 28.74 & 28.74 & 28.86  \\ 
        Koran & 15.04 & 16.18 & 16.98 & 13.37 & 14.02 & 14.16 & 15.64 & 17.00 & 14.21  \\ 
        Bible & 28.33 & 27.41 & 29.85 & 21.19 & 27.22 & 26.34 & 30.65 & 44.30 & 33.17  \\
        Covid & 38.08 & 32.34 & 32.45 & 29.92 & 31.68 & 30.85 & 34.58 & 33.49 & 33.68  \\ 
        ECB & 43.62 & 36.76 & 36.30 & 32.18 & 33.36 & 33.60 & 34.78 & 37.99 & 38.34  \\ 
        Globalvoices & 35.22 & 33.59 & 34.08 & 31.30 & 32.03 & 31.77 & 33.12 & 33.97 & 31.94  \\ 
        News(WMT14) & 39.68 & 36.68 & 37.58 & 33.13 & $\backslash$ & 33.77 & 36.29 & 37.00 & 33.80  \\
        Mixed(WMT22) & 33.29 & 33.64 & 33.97 & 29.72 & 29.09 & 30.00 & 29.58 & 33.07 & 26.02  \\ 
        Tedtalk14 & 38.60 & 36.81 & 37.21 & 33.07 & 34.46 & 33.53 & 36.06 & 36.54 & 35.60  \\
        Tedtalk17 & 32.61 & 31.22 & 31.96 & 28.44 & 29.05 & 28.75 & 30.54 & 30.25 & 30.56  \\ 
        TED2020 & 36.55 & 35.24 & 36.08 & 31.70 & 32.64 & 31.98 & 34.75 & 34.71 & 32.81  \\
        Chat(WMT20) & 52.48 & 51.46 & 52.85 & 44.98 & 46.68 & 46.56 & 48.49 & 49.50 & 47.21  \\
        Chat(WMT22) & 57.22 & 55.87 & 59.48 & 48.97 & 49.02 & 51.34 & 51.19 & 57.37 & 45.85  \\
        Medical(WMT14) & 52.39 & 46.56 & 46.04 & 41.00 & 43.24 & 42.66 & 44.98 & 47.30 & 45.11  \\ 
        Bio(WMT20) & 39.46 & 37.92 & 38.40 & 33.91 & 34.44 & 36.09 & 36.55 & 38.73 & 36.04  \\ 
        Bio(WMT22) & 43.38 & 41.63 & 41.61 & 36.55 & 36.77 & 38.38 & 38.19 & 41.72 & 36.48  \\ 
        IT(WMT16) & 58.62 & 54.97 & 55.45 & 40.29 & 42.68 & 48.18 & 51.83 & 55.20 & 51.56  \\ 
        Book & 22.20 & 20.33 & 20.85 & 17.54 & 19.37 & 20.07 & 20.44 & 24.22 & 20.54  \\ 
        RF & 30.49 & 28.93 & 30.26 & 28.05 & 29.77 & 33.44 & 29.07 & 32.22 & 28.56  \\ 
        Multiun & 54.54 & 46.04 & 46.03 & 42.48 & 44.38 & 45.12 & 47.17 & 54.35 & 49.83  \\ 
        Tilde & 38.92 & 30.67 & 31.03 & 28.24 & 29.53 & 30.11 & 32.55 & 34.39 & 35.39  \\ 
        QED & 39.05 & 37.60 & 38.23 & 33.41 & 34.79 & 34.53 & 36.31 & 36.90 & 35.77  \\ \hdashline
        Avg. & 39.85 & 36.61 & 37.11 & 32.29 & 33.61 & 34.18 & 35.71 & 38.41 & 36.12 \\ 
\bottomrule
\end{tabular}
}
\caption{Detailed BLEU scores of different models on the multi-domain test set for German to English translation.}
 \label{tab:details_de_en_bleu}
\end{table*}

\begin{table*}[htbp]
\centering
\resizebox{\textwidth}{!}{
\begin{tabular}{l|ccccccccc}
\toprule
        ~ & Google & ChatGPT & GPT4 & ParroT-7b & BayLing-7b & LLaMA-2-7b & ALMA-7b & LLaMA-2-70b & LLaMA-2-7b + FT  \\ \hline
        IT(OPUS) & 83.79 & 83.07 & 83.15 & 80.28 & 80.94 & 81.07 & 81.75 & 82.50 & 84.95  \\
        Medical & 85.03 & 83.98 & 84.01 & 82.55 & 82.81 & 82.77 & 83.44 & 83.21 & 84.52  \\ 
        Law & 85.92 & 84.71 & 84.77 & 82.33 & 82.72 & 82.28 & 83.23 & 84.46 & 85.50  \\ 
        Subtitles & 80.71 & 80.87 & 81.08 & 78.74 & 78.54 & 79.00 & 79.59 & 80.15 & 79.54  \\ 
        Koran & 72.59 & 73.85 & 74.57 & 70.91 & 70.72 & 71.37 & 72.31 & 72.83 & 70.52  \\ 
        Bible & 76.58 & 77.20 & 77.95 & 73.43 & 74.30 & 74.56 & 75.23 & 78.02 & 74.17  \\ 
        Covid & 87.72 & 87.32 & 87.50 & 86.30 & 86.78 & 86.53 & 87.21 & 87.01 & 86.71  \\ 
        ECB & 84.17 & 83.42 & 83.48 & 81.27 & 81.91 & 81.21 & 81.93 & 82.46 & 82.71  \\ 
        Globalvoices & 86.54 & 86.25 & 86.47 & 85.29 & 85.51 & 85.56 & 86.01 & 85.87 & 85.42  \\ 
        News(WMT14) & 86.96 & 86.90 & 87.17 & 85.24 & $\backslash$ & 85.43 & 86.24 & 86.30 & 85.07  \\ 
        Mixed(WMT22) & 85.15 & 85.53 & 85.64 & 83.26 & 83.26 & 83.47 & 84.14 & 84.61 & 82.48  \\ 
        Tedtalk14 & 86.62 & 86.56 & 86.64 & 84.82 & 84.86 & 85.36 & 85.95 & 86.11 & 85.18  \\ 
        Tedtalk17 & 86.36 & 85.98 & 86.24 & 84.65 & 84.78 & 84.92 & 85.79 & 85.67 & 84.58  \\ 
        TED2020 & 86.24 & 86.15 & 86.39 & 84.40 & 84.55 & 84.79 & 85.42 & 85.68 & 84.42  \\ 
        Chat(WMT20) & 92.07 & 91.73 & 92.19 & 90.43 & 90.57 & 90.86 & 91.47 & 91.27 & 90.97  \\
        Chat(WMT22) & 92.31 & 91.70 & 92.59 & 87.43 & 89.14 & 90.47 & 90.85 & 91.47 & 90.17  \\
        Medical(WMT14) & 89.64 & 89.17 & 89.12 & 87.61 & 87.94 & 88.02 & 88.44 & 88.97 & 88.23  \\ 
        Bio(WMT20) & 87.38 & 87.55 & 87.56 & 83.68 & 84.94 & 86.12 & 86.63 & 85.26 & 85.05  \\ 
        Bio(WMT22) & 85.85 & 86.27 & 86.27 & 84.63 & 84.77 & 84.51 & 85.00 & 85.42 & 84.75  \\
        IT(WMT16) & 91.55 & 91.47 & 91.55 & 87.94 & 88.34 & 89.82 & 90.54 & 90.86 & 90.36  \\ 
        Book & 78.60 & 78.56 & 78.96 & 76.45 & 76.92 & 77.27 & 76.91 & 78.18 & 76.68  \\ 
        RF & 85.48 & 85.28 & 85.54 & 84.52 & 84.78 & 85.19 & 85.03 & 85.07 & 84.70  \\ 
        Multiun & 87.62 & 86.81 & 87.12 & 85.22 & 85.72 & 85.46 & 86.02 & 87.19 & 86.52  \\
        Tilde & 87.57 & 86.66 & 86.88 & 85.30 & 85.62 & 85.68 & 86.27 & 86.48 & 86.34  \\ 
        QED & 87.08 & 86.85 & 87.12 & 85.34 & 85.27 & 85.55 & 86.01 & 86.36 & 85.62  \\ \hdashline
        Avg. & 85.58 & 85.35 & 85.60 & 83.28 & 83.57 & 83.89 & 84.46 & 84.86 & 84.21 \\ 
\bottomrule
\end{tabular}
}
\caption{Detailed COMET scores of different models on the multi-domain test set for German to English translation.}
 \label{tab:details_de_en_comet}
\end{table*}

\begin{table*}[htbp]
\centering
\resizebox{\textwidth}{!}{
\begin{tabular}{l|ccccccccc}
\toprule
        ~ & Google & ChatGPT & ParroT-7b & BayLing-7b & LLaMA-2-7b & ALMA-7b & LLaMA-2-70b & LLaMA-2-7b + FT  \\ \hline
        IT(OPUS) & 31.65 & 30.93 & 25.17 & 25.85 & 27.95 & 29.48 & 32.48 & 32.03  \\
        Medical & 42.09 & 37.32 & 29.76 & 30.67 & 30.94 & 35.15 & 35.03 & 38.04  \\
        Law & 41.91 & 33.06 & 21.86 & 23.57 & 21.79 & 28.58 & 31.40 & 36.61  \\
        Subtitles & 25.25 & 24.29 & 19.62 & 19.06 & 20.55 & 21.65 & 24.64 & 22.22  \\
        Koran & 11.40 & 12.41 & 7.74 & 8.37 & 9.30 & 10.32 & 11.11 & 9.69  \\
        Bible & 20.98 & 18.94 & 11.06 & 11.59 & 13.98 & 15.87 & 18.14 & 14.75  \\
        Covid & 32.36 & 26.99 & 21.89 & 21.25 & 20.82 & 25.09 & 25.92 & 22.06  \\
        ECB & 34.95 & 29.57 & 22.44 & 22.77 & 23.70 & 24.97 & 27.68 & 26.42  \\
        Globalvoices & 28.70 & 28.10 & 23.79 & 23.34 & 23.25 & 25.73 & 27.34 & 23.00  \\
        News(WMT14) & 35.04 & 29.35 & 25.09 & / & 23.73 & 28.38 & 29.31 & 23.99  \\
        Mixed(WMT22) & 37.52 & 33.40 & 26.15 & 25.63 & 24.51 & 30.23 & 31.73 & 25.50  \\
        Tedtalk14 & 31.33 & 30.58 & 24.90 & 24.66 & 25.54 & 27.63 & 30.58 & 25.80  \\
        Tedtalk17 & 27.97 & 26.77 & 22.37 & 22.48 & 22.78 & 25.01 & 25.99 & 23.83  \\
        TED2020 & 30.35 & 29.50 & 24.68 & 24.37 & 24.39 & 27.30 & 29.56 & 24.67  \\
        Chat(WMT20) & 50.38 & 44.74 & 32.80 & 33.72 & 30.65 & 38.58 & 37.24 & 32.66  \\
        Chat(WMT22) & 51.50 & 43.40 & 29.83 & 28.82 & 29.07 & 36.10 & 39.74 & 31.60  \\
        Medical(WMT14) & 34.99 & 32.91 & 25.37 & 24.91 & 25.10 & 28.16 & 31.31 & 25.83  \\
        Bio(WMT20) & 32.64 & 30.15 & 23.44 & 23.76 & 24.96 & 27.53 & 30.26 & 27.24  \\
        Bio(WMT22) & 37.72 & 33.25 & 25.99 & 26.67 & 25.79 & 29.91 & 31.79 & 28.02  \\
        IT(WMT16) & 42.55 & 41.71 & 31.25 & 32.53 & 32.61 & 39.02 & 43.46 & 36.48  \\
        Book & 15.74 & 16.32 & 11.74 & 12.84 & 13.32 & 13.01 & 16.06 & 14.18  \\
        RF & 26.05 & 24.78 & 21.26 & 20.97 & 19.91 & 22.29 & 25.01 & 20.75  \\
        Multiun & 42.73 & 35.04 & 25.58 & 27.61 & 26.80 & 30.02 & 33.94 & 30.03  \\
        Tilde & 31.41 & 26.24 & 19.94 & 19.46 & 19.62 & 22.94 & 25.04 & 22.99  \\
        QED & 30.79 & 30.59 & 25.02 & 25.25 & 25.95 & 27.84 & 30.74 & 25.16  \\ \hdashline
        Avg. & 33.12 & 30.01 & 23.15 & 23.34 & 23.48 & 26.83 & 29.02 & 25.74 \\
\bottomrule
\end{tabular}
}
\caption{Detailed BLEU scores of different models on the multi-domain test set for English to German translation.}
 \label{tab:details_en_de_bleu}
\end{table*}

\begin{table*}[htbp]
\centering
\resizebox{\textwidth}{!}{
\begin{tabular}{l|cccccccccc}
\toprule
        ~ & Google & ChatGPT & ParroT-7b & BayLing-7b & LLaMA-2-7b & ALMA-7b & LLaMA-2-70b & LLaMA-2-7b + FT  \\ \hline
        IT(OPUS) & 82.19 & 80.64 & 76.44 & 77.32 & 77.96 & 79.54 & 80.31 & 81.00  \\
        Medical & 84.29 & 83.01 & 79.38 & 79.87 & 80.54 & 82.14 & 81.76 & 83.16  \\
        Law & 86.74 & 85.12 & 79.66 & 80.94 & 77.33 & 83.62 & 83.74 & 85.19  \\
        Subtitles & 79.76 & 79.17 & 75.49 & 75.08 & 75.91 & 77.54 & 78.17 & 77.05  \\
        Koran & 73.44 & 73.25 & 67.66 & 68.02 & 69.54 & 71.61 & 71.61 & 70.41  \\
        Bible & 76.95 & 76.47 & 70.03 & 70.74 & 73.04 & 74.50 & 75.32 & 73.02  \\
        Covid & 88.26 & 87.38 & 84.59 & 84.63 & 84.50 & 86.69 & 86.49 & 85.43  \\
        ECB & 83.73 & 82.64 & 78.51 & 79.74 & 79.71 & 81.34 & 81.57 & 80.96  \\
        Globalvoices & 86.10 & 85.61 & 82.74 & 82.87 & 82.89 & 84.67 & 84.28 & 83.13  \\
        News(WMT14) & 88.53 & 87.36 & 84.40 & $/$ & 84.19 & 86.47 & 86.30 & 84.82  \\
        Mixed(WMT22) & 88.02 & 86.94 & 81.69 & 82.09 & 81.78 & 85.53 & 84.37 & 83.35  \\
        Tedtalk14 & 85.01 & 84.39 & 80.15 & 80.36 & 81.09 & 82.97 & 82.76 & 81.58  \\
        Tedtalk17 & 85.72 & 84.74 & 80.83 & 80.73 & 80.94 & 83.61 & 83.06 & 81.73  \\
        TED2020 & 85.30 & 84.52 & 80.50 & 80.74 & 81.02 & 83.17 & 82.97 & 81.52  \\
        Chat(WMT20) & 90.41 & 89.56 & 83.26 & 84.55 & 84.39 & 87.29 & 86.46 & 85.12  \\
        Chat(WMT22) & 90.44 & 88.14 & 83.60 & 84.30 & 84.93 & 87.28 & 86.51 & 85.35  \\
        Medical(WMT14) & 87.56 & 86.92 & 82.88 & 83.01 & 83.50 & 85.48 & 85.68 & 84.77  \\
        Bio(WMT20) & 87.44 & 87.05 & 79.88 & 80.67 & 82.12 & 82.43 & 82.94 & 85.41  \\
        Bio(WMT22) & 85.70 & 85.28 & 81.91 & 82.09 & 82.70 & 84.27 & 84.39 & 83.51  \\
        IT(WMT16) & 89.31 & 89.16 & 83.37 & 84.44 & 85.21 & 88.02 & 87.99 & 86.69  \\
        Book & 77.74 & 77.10 & 72.80 & 72.69 & 73.68 & 75.69 & 75.99 & 74.17  \\
        RF & 87.76 & 87.35 & 85.21 & 85.07 & 85.34 & 86.49 & 86.76 & 85.83  \\
        Multiun & 86.56 & 85.89 & 79.98 & 82.00 & 81.65 & 84.26 & 85.14 & 83.32  \\
        Tilde & 87.97 & 86.81 & 83.15 & 83.83 & 83.97 & 85.73 & 86.20 & 85.42  \\
        QED & 85.70 & 84.94 & 81.24 & 81.71 & 81.98 & 83.82 & 83.89 & 82.63  \\ \hdashline
        Avg. & 85.22 & 84.38 & 79.97 & 80.31 & 80.80 & 82.97 & 82.99 & 82.18 \\
\bottomrule
\end{tabular}
}
\caption{Detailed COMET scores of different models on the multi-domain test set for English to German translation.}
 \label{tab:details_en_de_comet}
\end{table*}

\begin{table*}[htbp]
\centering
\resizebox{\textwidth}{!}{
\begin{tabular}{l|cccccccccc}
\toprule
       ~ & Google & ChatGPT & ParroT-7b & BayLing-7b & LLaMA-2-7b & ALMA-7b & LLaMA-2-70b & LLaMA-2-7b + FT  \\ \hline
        News & 24.03 & 18.56 & 15.22 & 17.11 & 16.87 & 20.19 & 20.80 & 21.60  \\
        Science & 23.54 & 18.39 & 14.60 & 16.96 & 16.83 & 19.11 & 20.68 & 20.50  \\
        Law & 42.72 & 38.00 & 29.26 & 27.71 & 29.95 & 33.71 & 47.03 & 43.47  \\
        Subtitles & 19.86 & 17.96 & 13.49 & 15.61 & 15.21 & 17.24 & 18.20 & 18.34  \\
        Mixed(WMT22) & 29.24 & 26.23 & 20.20 & 20.61 & 21.53 & 23.46 & 25.61 & 18.13  \\
        News(WMT22) & 28.49 & 25.95 & 22.71 & 22.11 & 21.83 & 24.18 & 26.44 & 18.26  \\
        Social(WMT22) & 36.52 & 30.44 & 23.33 & 23.21 & 25.43 & 27.18 & 31.17 & 20.80  \\
        Conversation(WMT22) & 23.40 & 31.11 & 18.37 & 21.86 & 21.46 & 23.37 & 26.59 & 20.03  \\
        Ecommerce(WMT22) & 22.85 & 19.88 & 12.94 & 14.75 & 15.16 & 17.96 & 19.38 & 14.22  \\
        News(WMT19) & 37.48 & 33.28 & / & 27.11 & 27.46 & 34.01 & 33.42 & 21.96  \\
        Medical & 46.18 & 40.87 & 31.86 & 34.17 & 35.13 & 36.72 & 41.43 & 32.90  \\
        Bio(WMT20) & 31.94 & 27.46 & 21.09 & 23.41 & 23.32 & 26.34 & 28.53 & 24.27  \\
        Bio(WMT22) & 40.29 & 33.92 & 24.67 & 27.45 & 31.91 & 30.09 & 38.32 & 25.01  \\
        Tedtalk17 & 26.68 & 21.74 & 17.33 & 20.51 & 20.00 & 22.63 & 23.57 & 23.05  \\
        IT & 23.99 & 23.25 & 18.50 & 19.85 & 19.47 & 20.63 & 22.94 & 19.59  \\
        ITELEC & 43.34 & 35.99 & 29.07 & 31.78 & 31.13 & 33.36 & 37.79 & 30.50  \\
        Finance & 32.37 & 25.43 & 21.44 & 23.25 & 22.41 & 25.38 & 27.49 & 25.64  \\
        Car & 32.91 & 29.04 & 21.00 & 23.29 & 23.73 & 26.89 & 28.51 & 22.69  \\
        Machine & 35.92 & 30.41 & 21.95 & 23.72 & 24.07 & 26.88 & 29.89 & 22.90  \\
        Energyminera & 36.58 & 30.35 & 20.60 & 23.00 & 23.95 & 25.64 & 29.82 & 21.75  \\
        Webnovel1 & 29.19 & 22.66 & 12.24 & 15.04 & 18.99 & 18.62 & 23.61 & 13.84  \\
        Webnovel2 & 15.70 & 15.11 & 10.42 & 10.83 & 11.74 & 13.13 & 15.09 & 9.43  \\ \hdashline
        Avg. & 31.06 & 27.09 & 20.01 & 21.97 & 22.62 & 24.85 & 28.01 & 22.22 \\

\bottomrule
\end{tabular}
}
\caption{Detailed BLEU scores of different models on the multi-domain test set for Chinese to English translation.}
 \label{tab:details_zh_en_bleu}
\end{table*}

\begin{table*}[htbp]
\centering
\resizebox{\textwidth}{!}{
\begin{tabular}{l|cccccccccc}
\toprule
        ~ & Google & ChatGPT & ParroT-7b & BayLing-7b & LLaMA-2-7b & ALMA-7b & LLaMA-2-70b & LLaMA-2-7b + FT  \\ \hline
        News & 84.36 & 83.60 & 80.45 & 81.65 & 81.91 & 83.04 & 83.35 & 82.58  \\
        Science & 85.49 & 84.64 & 81.25 & 82.93 & 82.90 & 83.74 & 84.60 & 83.39  \\
        Law & 86.01 & 84.93 & 80.45 & 81.56 & 80.87 & 83.56 & 84.64 & 85.19  \\
        Subtitles & 81.63 & 80.85 & 76.08 & 78.41 & 78.66 & 79.64 & 80.29 & 78.97  \\
        Mixed(WMT22) & 81.07 & 82.80 & 75.85 & 77.72 & 77.93 & 79.86 & 80.33 & 76.83  \\
        News(WMT22) & 81.46 & 83.00 & 78.86 & 79.74 & 80.04 & 81.43 & 81.48 & 78.60  \\
        Social(WMT22) & 82.37 & 83.46 & 76.87 & 78.48 & 79.34 & 81.24 & 81.38 & 78.33  \\
        Conversation(WMT22) & 81.25 & 84.25 & 75.02 & 77.31 & 77.20 & 79.95 & 81.05 & 76.24  \\
        Ecommerce(WMT22) & 79.55 & 80.90 & 72.47 & 75.30 & 74.43 & 76.94 & 77.66 & 73.96  \\
        News(WMT19) & 83.27 & 84.18 & / & 80.66 & 81.25 & 83.30 & 83.05 & 80.15  \\
        Medical & 86.22 & 86.85 & 82.62 & 84.51 & 84.21 & 85.09 & 86.02 & 84.12  \\
        Bio(WMT20) & 83.87 & 83.89 & 79.88 & 82.01 & 80.70 & 82.54 & 82.70 & 80.42  \\
        Bio(WMT22) & 83.19 & 82.17 & 77.64 & 79.67 & 79.60 & 80.00 & 81.75 & 78.12  \\
        Tedtalk17 & 83.34 & 83.44 & 79.62 & 81.33 & 81.33 & 82.79 & 83.06 & 81.70  \\
        IT & 80.16 & 81.13 & 77.51 & 79.07 & 78.32 & 79.40 & 80.08 & 79.07  \\
        ITELEC & 86.83 & 86.53 & 82.49 & 84.30 & 83.38 & 84.92 & 85.78 & 83.70  \\
        Finance & 84.10 & 83.55 & 80.58 & 81.42 & 81.09 & 82.71 & 83.02 & 81.77  \\
        Car & 82.20 & 82.04 & 75.09 & 77.70 & 77.18 & 80.13 & 79.75 & 78.05  \\
        Machine & 84.77 & 84.25 & 79.20 & 80.95 & 80.42 & 82.21 & 83.04 & 81.05  \\
        Energyminera & 84.32 & 83.63 & 76.84 & 79.37 & 79.39 & 81.10 & 82.24 & 79.79  \\
        Webnovel1 & 77.83 & 77.95 & 69.15 & 70.98 & 72.85 & 74.97 & 76.30 & 71.40  \\
        Webnovel2 & 77.13 & 77.99 & 70.54 & 72.20 & 73.30 & 74.22 & 75.98 & 72.04  \\ \hdashline
        Avg. & 82.75 & 83.00 & 77.55 & 79.42 & 79.38 & 81.04 & 81.71 & 79.34 \\
\bottomrule
\end{tabular}
}
\caption{Detailed COMET scores of different models on the multi-domain test set for Chinese to English translation.}
 \label{tab:details_zh_en_comet}
\end{table*}

\begin{table*}[htbp]
\centering
\resizebox{\textwidth}{!}{
\begin{tabular}{l|cccccccccc}
\toprule
        ~ & Google & ChatGPT & ParroT-7b & BayLing-7b & LLaMA-2-7b & ALMA-7b & LLaMA-2-70b & LLaMA-2-7b + FT  \\ \hline
        News & 36.58 & 32.50 & 25.00 & 27.80 & 22.40 & 28.12 & 29.19 & 24.87  \\
        Science & 33.02 & 30.69 & 21.55 & 26.45 & 21.63 & 25.53 & 27.60 & 23.84  \\
        Law & 68.66 & 51.47 & 39.57 & 39.23 & 41.68 & 48.76 & 56.82 & 52.50  \\
        Subtitles & 26.32 & 24.69 & 18.40 & 22.19 & 18.47 & 22.13 & 23.74 & 20.50  \\
        Mixed(WMT22) & 50.58 & 43.91 & 30.40 & 36.99 & 29.25 & 36.64 & 38.54 & 32.17  \\
        News(WMT22) & 57.83 & 48.58 & 34.89 & 40.48 & 30.48 & 39.44 & 41.66 & 33.88  \\
        Social(WMT22) & 42.60 & 37.98 & 25.77 & 30.13 & 25.79 & 30.56 & 32.53 & 27.51  \\
        Conversation(WMT22) & 47.79 & 42.54 & 27.42 & 38.58 & 30.85 & 37.35 & 38.88 & 34.52  \\
        Ecommerce(WMT22) & 48.07 & 43.24 & 27.63 & 35.94 & 29.96 & 37.41 & 38.66 & 32.40  \\
        News(WMT19) & 56.42 & 43.69 & $/$ & 37.27 & 31.89 & 47.79 & 44.64 & 33.75  \\
        Medical & 56.56 & 49.74 & 27.42 & 38.48 & 31.39 & 41.20 & 43.96 & 32.44  \\
        Bio(WMT20) & 42.34 & 35.60 & 18.21 & 25.70 & 21.41 & 28.71 & 35.28 & 21.55  \\
        Bio(WMT22) & 49.58 & 40.28 & 25.98 & 33.78 & 26.39 & 34.32 & 40.20 & 28.03  \\
        Tedtalk17 & 31.72 & 30.90 & 22.92 & 26.12 & 23.43 & 26.52 & 28.86 & 24.23  \\
        IT & 28.12 & 25.46 & 19.52 & 23.24 & 24.22 & 25.92 & 26.16 & 23.94  \\
        ITELEC & 53.72 & 46.20 & 32.52 & 39.12 & 36.32 & 41.82 & 44.80 & 37.90  \\
        Finance & 47.89 & 42.70 & 30.10 & 34.71 & 25.86 & 34.54 & 36.79 & 28.48  \\
        Car & 40.87 & 35.38 & 16.86 & 23.35 & 21.49 & 30.83 & 31.18 & 22.85  \\
        Machine & 48.75 & 36.79 & 22.99 & 29.50 & 27.35 & 36.11 & 36.58 & 29.43  \\
        Energyminera & 48.39 & 38.32 & 18.82 & 26.31 & 24.85 & 32.47 & 36.03 & 27.28  \\
        Webnovel1 & 31.60 & 25.18 & 14.16 & 16.05 & 16.61 & 20.75 & 20.44 & 16.61  \\
        Webnovel2 & 25.34 & 20.32 & 11.87 & 14.81 & 14.28 & 17.88 & 18.38 & 14.52  \\ \hdashline
        Avg. & 44.22 & 37.55 & 24.38 & 30.28 & 26.18 & 32.95 & 35.04 & 28.33 \\
\bottomrule
\end{tabular}
}
\caption{Detailed BLEU scores of different models on the multi-domain test set for English to Chinese translation.}
 \label{tab:details_en_zh_bleu}
\end{table*}

\begin{table*}[htbp]
\centering
\resizebox{\textwidth}{!}{
\begin{tabular}{l|cccccccccc}
\toprule
        ~ & Google & ChatGPT & ParroT-7b & BayLing-7b & LLaMA-2-7b & ALMA-7b & LLaMA-2-70b & LLaMA-2-7b + FT  \\ \hline
        News & 86.20 & 84.90 & 79.73 & 82.55 & 80.30 & 83.76 & 83.44 & 81.94  \\
        Science & 86.90 & 86.09 & 79.91 & 83.05 & 81.39 & 84.23 & 84.46 & 83.05  \\
        Law & 92.47 & 90.34 & 85.97 & 87.01 & 87.77 & 89.32 & 90.67 & 90.28  \\
        Subtitles & 81.64 & 80.72 & 75.70 & 78.09 & 75.77 & 79.33 & 79.48 & 77.08  \\
        Mixed(WMT22) & 88.16 & 87.08 & 80.44 & 83.43 & 81.58 & 85.28 & 84.23 & 82.97  \\
        News(WMT22) & 89.07 & 87.37 & 82.43 & 84.52 & 81.85 & 85.77 & 84.50 & 83.51  \\
        Social(WMT22) & 84.37 & 84.10 & 76.91 & 79.20 & 77.93 & 81.31 & 80.78 & 78.93  \\
        Conversation(WMT22) & 90.02 & 88.90 & 81.91 & 85.74 & 84.07 & 87.92 & 87.01 & 85.54  \\
        Ecommerce(WMT22) & 89.26 & 88.03 & 80.56 & 84.37 & 83.00 & 86.24 & 85.10 & 84.04  \\
        News(WMT19) & 90.72 & 89.45 & $/$ & 86.67 & 85.43 & 89.39 & 88.85 & 86.56  \\
        Medical & 89.30 & 88.94 & 80.90 & 85.74 & 83.20 & 86.50 & 87.06 & 84.61  \\
        Bio(WMT20) & 86.36 & 83.90 & 78.10 & 82.62 & 80.13 & 84.18 & 83.07 & 81.86  \\
        Bio(WMT22) & 87.00 & 85.71 & 79.09 & 83.11 & 80.49 & 84.26 & 84.11 & 81.82  \\
        Tedtalk17 & 84.54 & 84.11 & 78.78 & 80.46 & 79.68 & 82.28 & 82.29 & 80.27  \\
        IT & 84.60 & 83.78 & 79.29 & 81.75 & 81.91 & 82.60 & 83.33 & 82.53  \\
        ITELEC & 91.04 & 89.93 & 83.74 & 87.33 & 86.89 & 88.37 & 89.00 & 87.52  \\
        Finance & 87.44 & 86.20 & 80.24 & 82.64 & 79.84 & 84.37 & 83.97 & 81.74  \\
        Car & 87.19 & 85.86 & 74.04 & 80.16 & 78.57 & 83.42 & 83.53 & 80.61  \\
        Machine & 90.26 & 88.64 & 81.44 & 84.87 & 84.06 & 87.07 & 87.28 & 85.35  \\
        Energyminera & 90.55 & 89.32 & 79.35 & 84.78 & 83.26 & 86.84 & 87.05 & 85.09  \\
        Webnovel1 & 84.11 & 82.51 & 72.61 & 75.39 & 76.54 & 79.83 & 78.86 & 77.03  \\
        Webnovel2 & 84.32 & 82.82 & 74.62 & 77.17 & 77.67 & 81.02 & 79.57 & 78.71  \\ \hdashline
        Avg. & 87.52 & 86.30 & 79.32 & 82.76 & 81.42 & 84.70 & 84.44 & 82.77 \\
\bottomrule
\end{tabular}
}
\caption{Detailed COMET scores of different models on the multi-domain test set for English to Chinese translation.}
 \label{tab:details_en_zh_comet}
\end{table*}

\end{CJK*}
\end{document}